\documentclass{article}

\usepackage{microtype}
\usepackage{graphicx}
\usepackage{subcaption}
\usepackage{booktabs}

\usepackage{hyperref}

\usepackage[accepted]{icml2026}

\usepackage{amsmath}
\usepackage{amssymb}
\usepackage{mathtools}
\usepackage{amsthm}

\usepackage{algorithm}
\usepackage{algorithmic}
\usepackage{fontawesome5}
\usepackage{xcolor}
\usepackage{enumitem}
\usepackage{wrapfig}

\usepackage[capitalize,noabbrev]{cleveref}

\theoremstyle{plain}

\theoremstyle{definition}

\theoremstyle{remark}

\usepackage{amsmath,amsfonts,bm}

\def\eqref#1{equation~\ref{#1}}

\def\1{\bm{1}}

\DeclareMathAlphabet{\mathsfit}{\encodingdefault}{\sfdefault}{m}{sl}
\SetMathAlphabet{\mathsfit}{bold}{\encodingdefault}{\sfdefault}{bx}{n}

\definecolor{MyRed}{HTML}{ED1C24}
\definecolor{MyYellow}{HTML}{FDEE10}

\icmltitlerunning{ReflFlow: Dynamic Specular Reconstruction}

\begin{document}

\twocolumn[
  \icmltitle{ReflFlow: Learning Geometry-Guided Ray Tracing for Dynamic Specular Reconstruction}

  \begin{icmlauthorlist}
    \icmlauthor{Jiachen Tao}{uic}
    \icmlauthor{Junyi Wu}{uic}
    \icmlauthor{Haoxuan Wang}{uic}
    \icmlauthor{Zongxin Yang}{harvard}
    \icmlauthor{Dawen Cai}{umich}
    \icmlauthor{Yan Yan}{uic}
  \end{icmlauthorlist}

  \icmlaffiliation{uic}{Department of Computer Science, University of Illinois Chicago, Chicago, IL, USA}
  \icmlaffiliation{harvard}{Harvard Medical School, Boston, MA, USA}
  \icmlaffiliation{umich}{University of Michigan, Ann Arbor, MI, USA}

  \icmlcorrespondingauthor{Yan Yan}{yyan55@uic.edu}

  \icmlkeywords{3D Reconstruction, Specular Scenes, Ray Tracing, Dynamic Scenes, Gaussian Splatting, Neural Rendering}

  \vskip 0.3in
]

\printAffiliationsAndNotice{}  %

\begin{abstract}
We present \textit{ReflFlow}, a novel framework for high-fidelity rendering of dynamic specular scenes by addressing two key challenges: precise reflection direction estimation and physically accurate modeling. To achieve this, we propose a Residual Material-Augmented 2D Gaussian Splatting representation that models dynamic geometry and material properties, allowing accurate reflection ray computation. Furthermore, we introduce a Dynamic Environment Gaussian and a hybrid rendering pipeline that decomposes rendering into diffuse and specular components, enabling physically informed specular synthesis via rasterization and ray tracing. Finally, we devise a coarse-to-fine training strategy to improve optimization stability and promote physically meaningful decomposition. Extensive experiments on dynamic scene benchmarks demonstrate that \textit{ReflFlow} outperforms prior methods quantitatively and qualitatively, producing sharper and more realistic specular reflections in complex dynamic environments.
\end{abstract}

\section{Introduction}\label{sec:introduction}
High-quality dynamic reconstruction and photorealistic rendering from monocular videos are essential for a wide range of applications, including augmented/virtual reality (AR/VR), 4D content creation, and artistic production. In recent years, Neural Radiance Fields (NeRF) \citep{nerf} and 3D Gaussian Splatting (3DGS) \citep{3dgs} have emerged as groundbreaking techniques in 3D reconstruction, also driving progress in monocular dynamic scene modeling. In particular, 3DGS represents a scene as a collection of 3D Gaussians and employs a rasterization-based rendering pipeline, greatly improving the efficiency of novel view synthesis. However, extending 3DGS to faithfully model dynamic scenes with specular surfaces remains challenging, primarily due to the difficulty of precise geometry estimation and ensuring physically accurate reflection modeling throughout the dynamic process.
\begin{figure*}[t!]
    \centering
    \includegraphics[width=\linewidth]{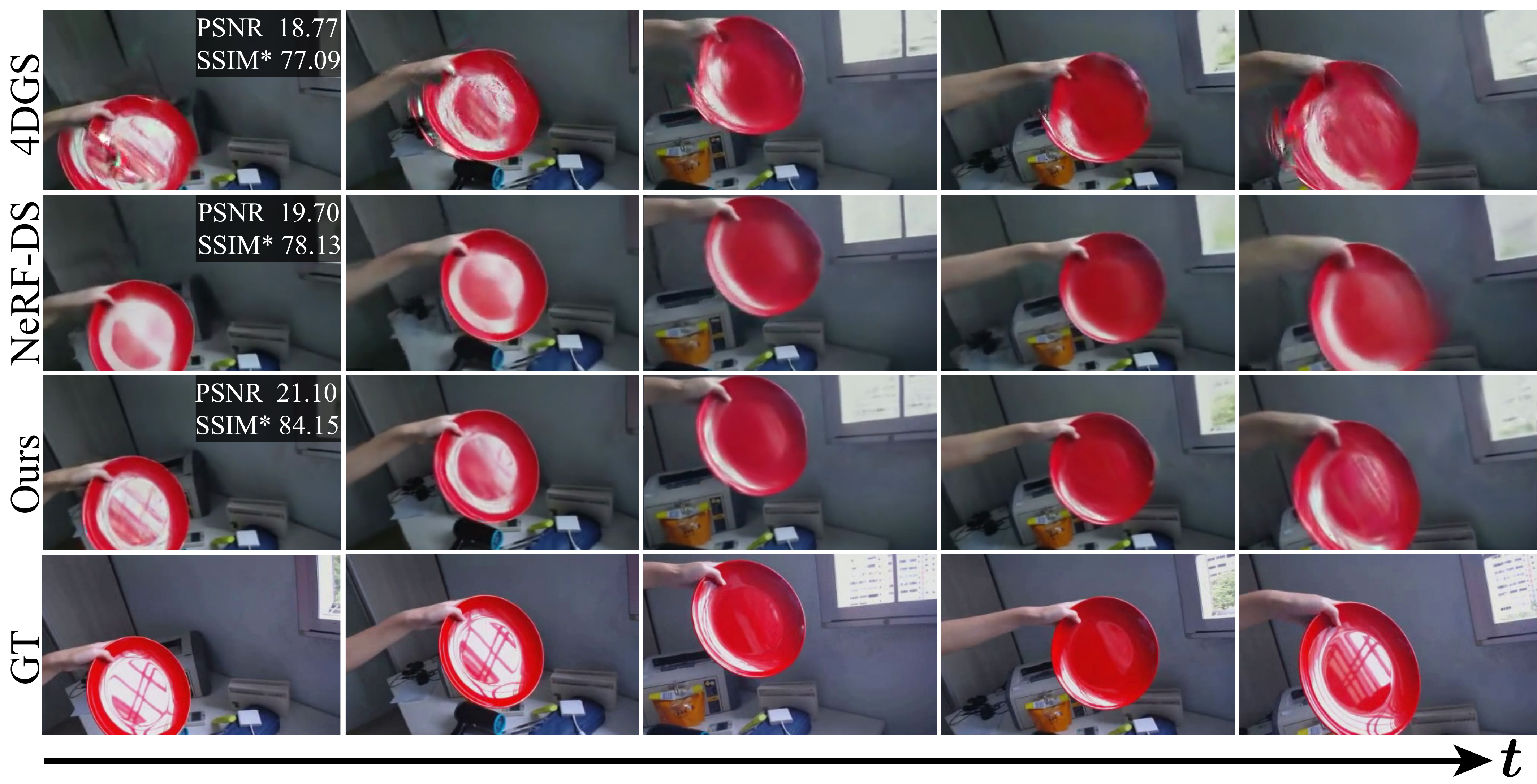}  %
    \vspace{-20pt}
    \caption{\textbf{ReflFlow} shows the sharpest and most photorealistic specular details among all compared approaches. PSNR~$\uparrow$ and SSIM~$\uparrow$ should be as high as possible. The performance shown in the figure corresponds to the \textit{Plate} scene. Please \faSearchPlus ~zoom in for a clearer view.}
    \label{fig:teaser}
\end{figure*}

Recently, several works have begun to consider view-dependent dynamic reconstruction. \citet{yan2023nerfdsneuralradiancefields} achieves dynamic view-dependent specular reconstruction by conditioning the radiance field on per-frame surface orientation in the observation space. To better capture view-dependent effects, \citet{gao20257dgsunifiedspatialtemporalangulargaussian} proposes a 7D Gaussian representation that incorporates spatial, temporal, and directional information. \citet{fan2024spectromotion} further advances this direction by dynamically decomposing rendering into diffuse and specular components and introducing a dynamic environment map, achieving improved modeling of dynamic specular reflections.

Physically, in dynamic specular reconstruction, specular details arise from the reflection of rays, which requires careful consideration of the reflection ray direction and simulation process of reflection. 
Recent view-dependent methods have introduced the use of reflection directions and have physically approximated the specular imaging process by employing dynamic environment maps: incident rays reflect off surfaces, and outgoing rays query the environment map to estimate the surface appearance. 

However, two key issues remain. 
\textbf{First}, 
the calculation of reflection ray directions is often highly approximate. 
Since 3DGS-based methods do not explicitly reconstruct surfaces, surface normals are typically estimated approximately. This approximation can cause deviations in reflection directions, which lead to inaccuracies in specular color computation. \textbf{Second}, while dynamic environment maps can approximate far-field reflections, they cannot accurately model near-field reflections and are limited by the resolution of the environment map, resulting in a loss of fine details.

In light of the preceding discussions, we present \textit{ReflFlow}, a novel framework for dynamic view-dependent reconstruction, explicitly designed to address the challenges in modeling complex specular reflections within dynamic scenes. ReflFlow comprises three key components: \textbf{First}, a Residual Material-Augmented 2D Gaussian Splatting representation that accurately captures dynamic geometry and temporally evolving material properties, ensuring precise reflection ray computation without normal estimation inaccuracies. \textbf{Second}, a Dynamic Environment Gaussian representation combined with a physically informed hybrid rendering pipeline, explicitly decomposing appearance into diffuse and specular components, enabling high-quality reconstruction of dynamic specular reflections. \textbf{Third}, a carefully designed coarse-to-fine training strategy stabilizes training and guides the model toward physically meaningful decomposition, resulting in robust and photorealistic novel view synthesis from monocular videos of dynamic specular scenes.

Our evaluations demonstrate that \textit{ReflFlow} achieves state-of-the-art performance on dynamic scene benchmarks with complex specular reflections. As shown in~\autoref{fig:teaser}, our method produces the sharpest and most photorealistic specular details among all compared approaches. Quantitatively, ReflFlow outperforms prior works across multiple metrics, achieving improvements of 0.74 in PSNR, 0.0358 in SSIM, and 0.0307 in LPIPS compared to the previous state-of-the-art, validating its effectiveness in dynamic specular reconstruction and photorealistic novel view synthesis.

\paragraph{Conflict of Interest Disclosure.}
The authors declare no financial conflicts of interest related to this work. All authors are affiliated with academic institutions, and no part of this research evaluates, endorses, or otherwise benefits any commercial product or service from a company employing any of the authors.

\section{Related Work}\label{sec:related}

\noindent\textbf{Specular Scene Reconstruction.} Neural Radiance Field (NeRF) \citep{nerf} and 3D Gaussian Splatting (3DGS) \citep{3dgs} have emerged as a significant advancement in computer graphics and 3D vision, achieving high-fidelity rendering quality. Numerous works have been proposed to improve rendering quality \citep{mipnerf,mipnerf360,zipnerf,yu2024mipsplatting,lu2024scaffoldgs,bi2024gs3,zhou2025dreamrenderer,wang2025xfieldphysicallygroundedrepresentation}, rendering efficiency \citep{liu2020neural,tensorf,plenoxels,mueller2022instant,DVGO,compact3dgs,bagdasarian2024zip,li2023human101}, geometry quality \citep{yariv2020multiview,wang2021neus,liu2023nero,wang2023neus2,li2023neuralangelo,wang2024unisdf,2dgs,chen2024pgsr,chen2024vcr,li2025tsgs}, and training optimization \citep{kheradmand2024mcmcgs,hollein2024lm}. However, these methods typically model specular effects either by directly encoding view direction or by relying on spherical harmonics (SH). Due to solely relying on viewing ray direction information, these methods often struggle to accurately capture high-frequency specular details, which frequently results in blurry reflections. 

To address this, mainstream approaches \citep{verbin2022refnerf,ma2023specnerf,jiang2023gaussianshader,liang2023envidr,verbin2024nerfcasting,tang2024threeigs,ye2024gsdr,chen2024pisr,xie2024envgs,gu2024IRGS,tao2026particlesfieldsreframingphoton} typically decompose rendering into diffuse and specular components. To capture specular reflections, one key is to utilize incident ray direction and outgoing ray direction information, either by using implicit neural networks \citep{verbin2022refnerf} to model lighting conditions or by leveraging explicit environment representations \citep{jiang2023gaussianshader,xie2024envgs} to improve reflection modeling capabilities. Another key is improving the quality of surface geometry and the accuracy of normal estimation \citep{ge2023refneus,liang2023envidr,liang2023spidr,liu2023nero,zhang2023neilfpp,chen2024pisr,yang2024specgaussian,zhu2024gsror,li2026transnormal}, which enables more precise reflection ray directions and thereby strengthens the modeling of reflective effects. Nevertheless, accurately and physically modeling dynamic environments and time-varying specular reflections remains a significant challenge. To address this, our work proposes a novel approach that incorporates a deformable environment representation along with additional explicit Gaussian attributes, specifically designed to capture temporal variations in specular color.

\noindent\textbf{Dynamic Scene Reconstruction.}
Recent advances in dynamic scene reconstruction have largely built upon two prominent paradigms: Neural Radiance Fields (NeRF) \citep{nerf} and 3D Gaussian Splatting (3DGS) \citep{3dgs}. \citet{nerf} revolutionized novel view synthesis by representing scenes as continuous volumetric functions parameterized by neural networks. While initially designed for static scenes, a range of extensions \citep{pumarola2020dnerf,li2021nsff,park2021nerfies,park2021hypernerf,tretschk2021nonrigid,xian2021stnerf,guo2023forwardflow,liu2023robustdynrf,chen2024narcan,ma2024humannerfse,wu2025denver} have adapted NeRFs for dynamic scenarios. These include D-NeRF \citep{pumarola2020dnerf}, Nerfies \citep{park2021nerfies}, and HyperNeRF \citep{park2021hypernerf}, which condition on time and learn deformation fields to warp points across timesteps. Other methods, such as DyNeRF \citep{liu2023robustdynrf}, use compact latent codes for time-conditioned radiance fields, and HexPlane \citep{Cao2023HEXPLANE} accelerates rendering via hybrid representations. Despite these efforts, NeRF-based approaches remain computationally intensive and often struggle with real-time performance and accurate modeling of view-dependent effects in complex dynamic scenes.

To address these challenges, 3D Gaussian Splatting \citep{3dgs} has emerged as a promising alternative, offering high-quality, real-time rendering via rasterization of 3D Gaussians with learnable parameters. Building on this foundation, several works \citep{liang2023gaufre,wu2023fourdgs,yang2023deformablegs,huang2024scgs,stearns2024marbles,wang2024shapeofmotion,yang2023gs4d,gao20246dgs,motiongs,gao20257dgsunifiedspatialtemporalangulargaussian} have extended 3DGS to dynamic settings. Some methods \citep{liang2023gaufre,wu2023fourdgs,yang2023deformablegs,huang2024scgs,stearns2024marbles,wang2024shapeofmotion} utilize deformable networks to add a residual component to the attributes of 3D Gaussians, embedding both temporal and spatial information into the representation. Other approaches \citep{yang2023gs4d,gao20246dgs,gao20257dgsunifiedspatialtemporalangulargaussian} extend 3DGS to higher-dimensional Gaussian distributions, treating the 3D Gaussians at each timestamp as a conditional distribution conditioned on time. More recently, \citet{fan2024spectromotion} introduced a dynamic environment map into dynamic scene reconstruction, enabling improved modeling of dynamic specular reflections. However, these methods still lack precise reflection direction estimation and physically accurate reflection modeling throughout the dynamic process. To address these limitations, our work proposes a new approach that computes reflection ray directions without approximation and explicitly models the dynamic specular reflection process in a physically informed manner, thereby enabling accurate and temporally consistent reconstruction of complex dynamic specular effects.

\section{Preliminary}\label{sec:preliminary}

\noindent\textbf{2D Gaussian Splatting.} Our reconstruction stage builds upon the state-of-the-art point-based renderer with high-quality geometry performance, 2DGS \citep{2dgs}. 2DGS comprises several components: the central point $\mathbf{p}_k$, two principal tangential vectors $\mathbf{t}_u$ and $\mathbf{t}_v$ that determine its orientation, and a scaling vector $\mathbf{S} = (s_u, s_v)$ controlling the variances of the 2D Gaussian distribution. 
2D Gaussian Splatting represents the scene's geometry as a set of 2D Gaussians. A 2D Gaussian is defined in a local tangent plane in world space, parameterized as follows:
\begin{align}
P(u,v) &= \mathbf{p}_k + s_u \mathbf{t}_u u + s_v \mathbf{t}_v v .
\label{eq:plane-to-world}
\end{align}
For the point $\mathbf{u}=(u,v)$ in $uv$ space, its 2D Gaussian value can then be evaluated by:
\begin{equation}
\mathcal{G}(\mathbf{u}) = \exp\left(-\frac{u^2+v^2}{2}\right).
\label{eq:gaussian-2d}
\end{equation}
The center $\mathbf{p}_k$, scaling $(s_u, s_v)$, and the rotation $(\mathbf{t}_u, \mathbf{t}_v)$ are learnable parameters. 
Each 2D Gaussian primitive has opacity $\alpha$ and view-dependent appearance $\mathbf{c}$ with spherical harmonics. 

\begin{figure*}[t!]
    \centering
    \includegraphics[width=\linewidth]{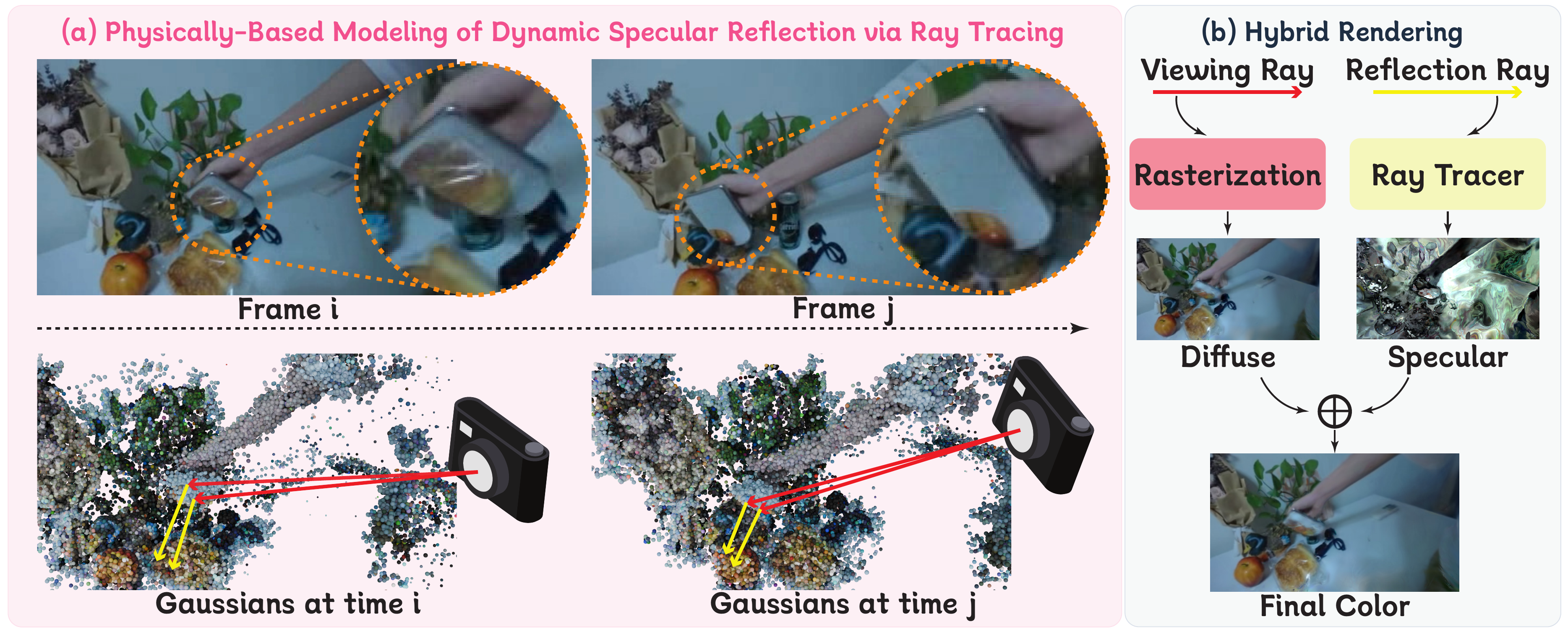}  %
    \caption{\textbf{Overview of ReflFlow.} (a) For a dynamic specular scene, at each timestamp, a \textbf{\textcolor{MyRed}{viewing ray}} is traced from the camera. After intersecting with the main content, it reflects off the surface based on the surface normal. The resulting \textbf{\textcolor{MyYellow}{reflection ray}} then intersects with the dynamic environment. (b) To render such a scene, we use rasterization to compute the diffuse color of the main content and employ a ray tracer to compute the specular color via the reflection ray. Finally, the diffuse and specular components are blended to the final color.}
    \label{fig:pipeline}
\end{figure*}
For volume rendering, Gaussians are sorted according to their depth value and composed into an image with front-to-back alpha blending:
\begin{equation}
\mathbf{c}(\mathbf{x}) = \sum_{i=1} \mathbf{c}_i \alpha_i \mathcal{G}_i(\mathbf{u}(\mathbf{x})) \prod_{j=1}^{i-1} (1 - \alpha_j \mathcal{G}_j(\mathbf{u}(\mathbf{x}))).
\end{equation}
where $\mathbf{x}$ represents a homogeneous ray emitted from the camera and passing through $uv$ space.

Compared to a 3DGS \citep{3dgs}, 2DGS \citep{2dgs} offers distinct advantages as a surface representation. First, the ray-splat intersection method adopted by 2DGS avoids multi-view depth inconsistency. Second, 2D Gaussians inherently provide a well-defined normal, which is defined by two orthogonal tangential vectors \( \mathbf{t}_w = \mathbf{t}_u \times \mathbf{t}_v \), thus avoiding approximations when computing surface normals and reflection ray directions, which is critical for capturing high-frequency specular details. However, 2DGS relies on the limited representational capacity of Spherical Harmonics (SH) to model view-dependent scene appearance and struggles to reconstruct dynamic scenes. To this end, we extend the geometry-aligned 2D Gaussian primitives to Residual Material-Augmented 2DGS and demonstrate how we effectively model complex dynamic reflections in the next section.

\section{Method}\label{sec:method}
\noindent\textbf{Overview of the approach.} Given a monocular video of a dynamic specular scene, our goal is to reconstruct the dynamic scene and synthesize photorealistic novel views in real-time. To ensure the quality of the dynamic scene geometry and the accuracy of reflection ray direction computation, as well as to effectively model material properties across different parts of the dynamic scene, we propose Residual Material-Augmented 2DGS to represent the dominant content of the dynamic scene. Building on this, we propose a Dynamic Environment Gaussian to learn the dynamic environment, enabling the computation of specular color through reflection rays in a physically informed manner. Finally, to further improve training stability, we propose a coarse-to-fine training strategy.

\subsection{Residual Material-Augmented 2DGS}

\noindent\textbf{Challenges in Normal Estimation for 3D Representation.} Normal estimation is critical for modeling specular because accurately determining the reflection ray direction relies on obtaining the surface normal $\mathbf{n}$. The reflection ray direction $\mathbf{d}_{\text{out}}$ is computed as follows, $\mathbf{d}_{\text{in}}$ is the incident ray direction:
\begin{equation}
\mathbf{d}_{\text{out}} = \mathbf{d}_{\text{in}} - 2 (\mathbf{d}_{\text{in}} \cdot \mathbf{n}) \mathbf{n}.
\end{equation}

However, accurate normal estimation on Gaussian spheres remains challenging. Although recent works~\citep{jiang2023gaussianshader,fan2024spectromotion} have proposed approximation-based methods for estimating normals, such approximations inevitably introduce errors. These errors propagate into computation of the reflection ray direction $\mathbf{d}_{\text{out}}$, further amplifying inaccuracies. As a result, fine details in specular effects may be significantly distorted or incorrectly reconstructed. This motivates the search for a representation that enables accurate and error-free normal computation. As discussed earlier in the preliminary section, 2DGS~\citep{2dgs} inherently provides well-defined normals without approximation during computation. However, 2DGS~\citep{2dgs} is originally designed for static scenes, struggles with dynamic reconstruction, and lacks ability to model surface material properties, which are essential for physically-based rendering (PBR)~\citep{pbr}.

\noindent\textbf{Residual Material-Augmented 2DGS.} Specular tint $\mathbf{s}_{\text{tint}} \in [0, 1]$~\citep{burley2012disney} is a key material property in physically based rendering (PBR)~\citep{pbr} frameworks. Specular tint controls the color of specular reflections based on the material’s intrinsic color. 
Accurately modeling these properties is essential for faithfully reproducing realistic appearance under varying lighting conditions. To capture the material properties of the 3D scene, we introduce $\mathbf{s}_{\text{tint}}$ as learnable parameters for each 2D Gaussian.

To enable the representation to capture time-varying information, we propose a Time-Conditioned Residual Network with parameters $\theta$ to predict a residual $\Delta \mathbf{G}^t=\{\Delta \mathbf{p}^t, \Delta \mathbf{s}^t, \Delta \mathbf{r}^t, \Delta \mathbf{o}^t, \Delta \mathbf{s}_\text{tint}^t \}$ that refines the parameters of the representation, where $\mathbf{G}$ denotes the Residual Material-Augmented 2DGS. The input to this network consists of the center position of each Gaussian $\mathbf{p}$ and the time $\mathbf{t}$:
\begin{equation}
\Delta \mathbf{G}^t = \mathcal{F}_{\theta_G}(\mathbf{p}, \mathbf{t}), \mathbf{p} \in \mathbb{R}^3, \mathbf{t} \in [0, 1]
\end{equation}
So that the deformed Gaussians \(\mathbf{G}^t\) at time \(t\) is obtained by $(\mathbf{p}^t, \mathbf{s}^t, \mathbf{r}^t, \mathbf{o}^t, \mathbf{s}_\text{tint}^t)
= (\Delta \mathbf{p}^t, \Delta \mathbf{s}^t, \Delta \mathbf{r}^t, \Delta \mathbf{o}^t, \Delta \mathbf{s}_\text{tint}^t)
+ (\mathbf{p}, \mathbf{s}, \mathbf{r}, \mathbf{o}, \mathbf{s}_\text{tint})$. To further improve the quality of the reconstructed geometry, we introduce additional supervision on the surface normals.

\noindent\textbf{Geometry-Aligned Normal Loss.}
Following 2DGS~\citep{2dgs}, we adopt a normal consistency loss $\mathcal{L}_{\text{norm}}$ to enforce consistency between the rendered normal map $\mathbf{n}$ and pseudo normal map $\mathbf{N}_d$ derived from the depth map. The pseudo normal map is computed via normalized cross-products of spatial depth gradients. The consistency loss is defined as:
\begin{equation}
\mathcal{L}_{\text{norm}} = \frac{1}{N_p} \sum_{i=1}^{N_p} \left(1 - \mathbf{n}_i^\top \mathbf{N}_d(\mathbf{u}_i)\right),
\end{equation}
where $N_p$ is the number of pixels, $\mathbf{n}_i$ is the predicted normal at pixel $i$, and $\mathbf{N}_d(\mathbf{u}_i)$ is the pseudo normal at pixel $\mathbf{u}_i$, computed as:
\begin{equation}
\mathbf{N}_d(\mathbf{u}) = \frac{\nabla_u \mathbf{P}_d \times \nabla_v \mathbf{P}_d}{\left\|\nabla_u \mathbf{P}_d \times \nabla_v \mathbf{P}_d\right\|},
\end{equation}
\noindent\textbf{Temporal-Consistent Normal Supervision Loss.}  
While $\mathcal{L}_{\text{norm}}$ provides a self-supervised constraint based on geometric consistency, it is often insufficient for supervising complex dynamic surfaces in the absence of explicit normal supervision. To overcome this limitation, we introduce a supervised loss $\mathcal{L}_{\text{tc-norm}}$ using normals $\mathbf{N}_e$ estimated by NormalCrafter~\citep{normalcrafter}, which leverages video diffusion priors to produce temporally consistent surface normals. Compared to other monocular normal estimators, this prior provides improved temporal consistency, effectively reducing frame-to-frame flickering and making it well-suited for supervising dynamic geometry in view-dependent scenarios.
\begin{equation}
\mathcal{L}_{\text{tc-norm}} = \frac{1}{N_p} \sum_{i=1}^{N_p} \left( 1 - \mathbf{n}_i^\top \mathbf{N}_e \right).
\end{equation}
\noindent\textbf{Summary.}
This approach captures dynamic motion while preserving high-quality geometry, allowing accurate reflection ray direction computation for dynamic scenes, which is an essential prerequisite for the subsequent physically based modeling of dynamic specular reflection.

\subsection{Physically Based Modeling of Dynamic Specular Reflection}
Given a reliable representation of the main content from Residual Material-Augmented 2DGS, the next critical step is to accurately model the reflection process. Specifically, incident rays intersect with the main object, reflect off its surface based on the surface normals, and subsequently intersect with the surrounding environment to determine the reflected illumination.

\noindent\textbf{Dynamic Environment Gaussian.} 
Recent methods~\citep{jiang2023gaussianshader,fan2024spectromotion} typically utilize dynamic environment maps to model dynamic illumination conditions. However, due to inherent limitations, environment maps often struggle to capture high-quality specular details. First, environment maps have limited resolution, resulting in blurred or insufficiently sharp specular reflections. Second, environment maps inherently assume distant illumination, failing to model near-field reflections, which are crucial for realistic rendering of close-proximity interactions.

To address these limitations, inspired by~\citep{xie2024envgs}, we introduce Dynamic Environment Gaussian representations $\mathbf{G}_{\text{env}}$ to model the dynamic environment. Each Gaussian in $\mathbf{G}_{\text{env}}$ is parameterized similarly to 2D Gaussian Splatting (2DGS), including attributes such as position $\mathbf{p}$, scale $\mathbf{s}$, rotation $\mathbf{r}$, and opacity $\mathbf{o}$. To capture temporal variations, we introduce a residual correction network $\mathcal{F}_{\theta_{\text{env}}}$ that predicts time-dependent residuals. Specifically, at timestamp $t$, the dynamic environment Gaussian $\mathbf{G}_{\text{env}}^t$ is defined by applying the residual corrections predicted by $\mathcal{F}_{\theta_{\text{env}}}$:
\begin{equation}
\Delta \mathbf{G}_{\text{env}}^t = \mathcal{F}_{\theta_{\text{env}}}(\mathbf{p}, \mathbf{t}), \quad \mathbf{p} \in \mathbb{R}^3, \mathbf{t} \in [0, 1],
\end{equation}
and the parameters at time $t$ are updated as:
\begin{equation}
\mathbf{G}_{\text{env}}^t = (\mathbf{p}, \mathbf{s}, \mathbf{r}, \mathbf{o}) + (\Delta \mathbf{p}^t, \Delta \mathbf{s}^t, \Delta \mathbf{r}^t, \Delta \mathbf{o}^t).
\end{equation}
This enables accurate modeling of time-varying environmental illumination and reflection dynamics.

\noindent\textbf{Color Decomposition.} Following the principles of physically based rendering (PBR)~\citep{pbr} and recent works~\citep{jiang2023gaussianshader,fan2024spectromotion,xie2024envgs}, we explicitly decompose the rendered color into diffuse $\mathbf{C}_{\text{diffuse}}$ and specular $\mathbf{C}_{\text{specular}}$ components. Such decomposition allows us to separately handle view-independent illumination (diffuse), primarily influenced by surface albedo and environmental lighting, and view-dependent illumination (specular), which depends on reflection directions and surface properties. This explicit separation enhances the accuracy and realism of specular reflections, enabling detailed control and modeling of complex reflective behaviors. Formally, the final rendered color $\mathbf{C}$ at each pixel is computed as:
\begin{equation}
\mathbf{C} = (1 - \alpha_{\text{spec}})\mathbf{C}_{\text{diffuse}} + \alpha_{\text{spec}}\mathbf{C}_{\text{specular}},
\end{equation}
where the blending weight $\alpha_{\text{spec}}$ balances the contribution between diffuse and specular components.

To derive $\alpha_{\text{spec}}$ from the material properties, we employ a separate rasterization process where each Gaussian contributes via its opacity-weighted specular tint $\mathbf{s}_{\text{tint}}$. This ensures that the specular blending weight is computed in a view-dependent manner through a transmittance-weighted sum over visible Gaussians:
\begin{equation}
\alpha_{\text{spec}} = \sum_{i \in \mathcal{N}} \mathbf{s}_{\text{tint},i} \alpha_i \prod_{j=1}^{i-1}(1 - \alpha_j),
\end{equation}
where $\mathbf{s}_{\text{tint},i}$ is the specular tint of the $i$-th Gaussian, and $\alpha_i$ is computed from a 2D Gaussian projection scaled by a learned per-point opacity. This formulation ensures that specular contribution is view-dependent and geometry-aware.

\noindent\textbf{Hybrid Rendering Pipeline.}
To efficiently and accurately synthesize view-dependent reflections, we employ a hybrid rendering pipeline that combines rasterization and physically-based ray tracing. Specifically, we first utilize the rasterization-based rendering pipeline provided by~\citep{2dgs} to compute the diffuse color $\mathbf{C}_{\text{diffuse}}$ using incident rays:
\begin{equation}
\mathbf{C}_{\text{diffuse}} = \sum_{i \in \mathcal{N}} \mathbf{c}_i \alpha_i \prod_{j=1}^{i-1}(1-\alpha_j),
\end{equation}
where $\mathbf{c}_i$ denotes the diffuse color attribute of the $i$-th Gaussian intersected by the ray, $\alpha_i$ is its opacity, and $\mathcal{N}$ represents the set of Gaussians along the ray.

Subsequently, we employ a physically informed ray tracer~\citep{xie2024envgs} to compute the specular color $\mathbf{C}_{\text{specular}}$ by tracing reflection rays guided by accurate surface normals. These rays query the Dynamic Environment Gaussian representation, modeling time-varying environment illumination. For each reflected ray, we collect up to $k$ Gaussian intersections and aggregate their contributions by spatial proximity and accumulated transmittance. The specular color $\mathbf{C}{\text{specular}}$ is computed as:
\begin{equation}
    \mathbf{C}_{\text{specular}} = \sum_{i=1}^{k} T_i \cdot \mathcal{G}_i(\mathbf{H}_i^{-1} \mathbf{x}_i) \cdot \mathbf{c}_i,
\end{equation}
where $\mathbf{x}_i$ is the intersection point between the reflection ray and the $i$-th Gaussian, $\mathbf{H}_i$ is its affine transformation matrix, $\mathbf{c}_i$ is the specular color attribute of the Gaussian, and $\mathcal{G}_i(\cdot)$ denotes the isotropic Gaussian kernel evaluated in the local coordinate system. $T_i = \prod_{j=1}^{i-1}(1 - \alpha_j)$ represents the accumulated transmittance along the ray, with $\alpha_j$ being the opacity of the $j$-th Gaussian.

\begin{figure*}[t!]
    \centering
    \includegraphics[width=\linewidth]{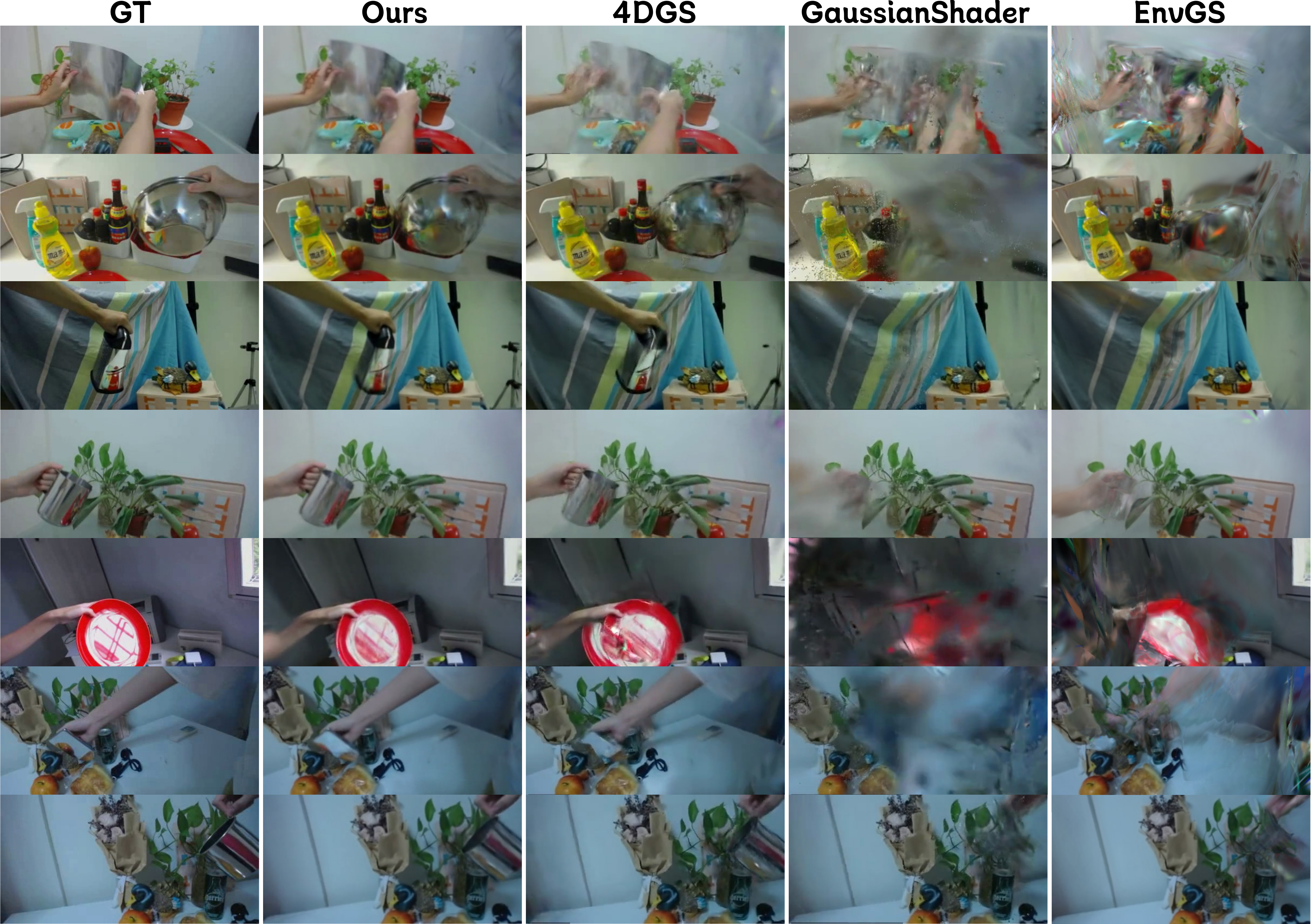}  %
    \caption{\textbf{Qualitative Comparison Results on the NeRF-DS Dataset.} Our method significantly improves the visual quality of dynamic specular reconstruction compared to previous approaches. In particular, it produces sharper details and fewer artifacts in specular regions, demonstrating enhanced fidelity in modeling dynamic reflections. Please \faSearchPlus ~zoom in for more details.}
    \label{fig:maincompare}
\end{figure*}
\noindent\textbf{Summary.} By explicitly modeling dynamic environments, decomposing appearance into diffuse and specular components, and combining rasterization with ray tracing, our framework achieves physically accurate reconstruction of dynamic specular effects.  
To ensure robust and stable convergence, we then introduce a coarse-to-fine training strategy tailored for dynamic scenes.

\subsection{Coarse-to-Fine Training Strategy}
\label{subsec:c2f}

Although our method explicitly decomposes the final color into diffuse and specular components, supervision is only applied to the final rendered color $\mathbf{C}$. As a result, the network receives no direct supervision for either $\mathbf{C}_{\text{diffuse}}$ or $\mathbf{C}_{\text{specular}}$, which makes the decomposition problem inherently ill-posed and potentially unstable, especially in the early stages of training. Without proper regularization, the network may converge to degenerate solutions that satisfy the color loss but fail to accurately separate physically meaningful reflectance components.

We begin training with the diffuse rendering branch only, focusing on reconstructing geometry and diffuse color from incident rays. This provides a stable geometric and photometric foundation for the network. Once the diffuse reconstruction reaches a reasonable quality, we progressively introduce the specular rendering branch and train the full model, allowing the ray-traced reflection components to learn the specular detail. Details of the strategy are provided in the supplementary material.

This staged training procedure improves convergence stability, reduces entanglement between diffuse and specular components, and promotes better geometry-material separation. It is particularly effective when learning from real-world monocular videos with complex specular effects.

\section{Experiments}\label{sec:exp}

\begin{table*}[t]
    \setlength{\tabcolsep}{3pt}
    \centering
    \small
    \caption{\textbf{Quantitative comparison on the NeRF-DS~\cite{yan2023nerfdsneuralradiancefields} dataset.} We report the average PSNR, SSIM, and LPIPS (VGG) across seven scenes. The \colorbox{red!25}{best}, the \colorbox{orange!25}{second best}, and the \colorbox{yellow!25}{third best} results are denoted by red, orange, yellow. }
    \label{tab:whole_scene_tab}
    \resizebox{\textwidth}{!}{%
    \begin{tabular}{lccccccccccccccc}
    \toprule
     & \multicolumn{3}{c}{As} & & \multicolumn{3}{c}{Basin} & & \multicolumn{3}{c}{Bell} & & \multicolumn{3}{c}{Cup} \\ 
    \cmidrule{2-4} \cmidrule{6-8} \cmidrule{10-12} \cmidrule{14-16}
    Method & PSNR↑ & SSIM↑ & LPIPS↓ & & PSNR↑ & SSIM↑ & LPIPS↓ & & PSNR↑ & SSIM↑ & LPIPS↓ & & PSNR↑ & SSIM↑ & LPIPS↓ \\ 
    \midrule
    Deformable 3DGS~\cite{yang2023deformablegs}
    & \colorbox{yellow!25}{26.04} & \colorbox{yellow!25}{0.8805} & \colorbox{yellow!25}{0.1850}
    & & 19.53 & 0.7855 & \colorbox{yellow!25}{0.1924}
    & & \colorbox{yellow!25}{23.96} & 0.7945 & 0.2767
    & & 24.49 & \colorbox{yellow!25}{0.8822} & \colorbox{yellow!25}{0.1658} \\

    4DGS~\cite{yang2023gs4d}
    & 24.85 & 0.8632 & 0.2038
    & & 19.26 & 0.7670 & 0.2196
    & & 22.86 & 0.8015 & \colorbox{yellow!25}{0.2061}
    & & 23.82 & 0.8695 & 0.1792 \\

    GaussianShader~\cite{jiang2023gaussianshader}
    & 21.89 & 0.7739 & 0.3620
    & & 17.79 & 0.6670 & 0.4187
    & & 20.69 & \colorbox{yellow!25}{0.8169} & 0.3024
    & & 20.40 & 0.7437 & 0.3385 \\

    GS-IR~\cite{liang2023gs}
    & 21.58 & 0.8033 & 0.3033
    & & 18.06 & 0.7248 & 0.3135
    & & 20.66 & 0.7829 & 0.2603
    & & 20.34 & 0.8193 & 0.2719 \\

    NeRF-DS~\cite{yan2023nerfdsneuralradiancefields}
    & 25.34 & 0.8803 & 0.2150
    & & \colorbox{yellow!25}{20.23} & \colorbox{yellow!25}{0.8053} & 0.2508
    & & 22.57 & 0.7811 & 0.2921
    & & \colorbox{yellow!25}{24.51} & 0.8802 & 0.1707 \\

    HyperNeRF~\cite{park2021hypernerf}
    & 17.59 & 0.8518 & 0.2390
    & & \colorbox{red!25}{22.58} & \colorbox{orange!25}{0.8156} & 0.2497
    & & 19.80 & 0.7650 & 0.2999
    & & 15.45 & 0.8295 & 0.2302 \\

    EnvGS~\cite{xie2024envgs}
    & 21.59 & 0.7925 & 0.2997
    & & 17.95 & 0.7506 & 0.2855
    & & 20.75 & 0.7998 & 0.2331
    & & 20.25 & 0.8074 & 0.2717 \\

    SpectroMotion~\cite{wang2024shapeofmotion}
    & \colorbox{red!25}{26.80} & \colorbox{orange!25}{0.8843} & \colorbox{orange!25}{0.1761}
    & & 19.75 & 0.7915 & \colorbox{orange!25}{0.1896}
    & & \colorbox{orange!25}{25.46} & \colorbox{orange!25}{0.8490} & \colorbox{orange!25}{0.1600}
    & & \colorbox{orange!25}{24.65} & \colorbox{orange!25}{0.8871} & \colorbox{orange!25}{0.1588} \\

    Ours
    & \colorbox{orange!25}{26.73} & \colorbox{red!25}{0.9026} & \colorbox{red!25}{0.1560}
    & & \colorbox{orange!25}{20.42} & \colorbox{red!25}{0.8479} & \colorbox{red!25}{0.1514}
    & & \colorbox{red!25}{25.69} & \colorbox{red!25}{0.8825} & \colorbox{red!25}{0.1205}
    & & \colorbox{red!25}{25.08} & \colorbox{red!25}{0.9082} & \colorbox{red!25}{0.1394} \\

    \midrule
     & \multicolumn{3}{c}{Plate} & & \multicolumn{3}{c}{Press} & & \multicolumn{3}{c}{Sieve} & & \multicolumn{3}{c}{\textbf{Mean}} \\
    \cmidrule{2-4} \cmidrule{6-8} \cmidrule{10-12} \cmidrule{14-16}
    Method & PSNR↑ & SSIM↑ & LPIPS↓ & & PSNR↑ & SSIM↑ & LPIPS↓ & & PSNR↑ & SSIM↑ & LPIPS↓ & & PSNR↑ & SSIM↑ & LPIPS↓ \\ 
    \midrule
    Deformable 3DGS~\cite{yang2023deformablegs}
    & 19.07 & 0.7352 & 0.3599
    & & \colorbox{yellow!25}{25.52} & 0.8594 & \colorbox{yellow!25}{0.1964}
    & & \colorbox{orange!25}{25.37} & 0.8616 & \colorbox{yellow!25}{0.1643}
    & & \colorbox{yellow!25}{23.43} & 0.8284 & 0.2201 \\

    4DGS~\cite{yang2023gs4d}
    & 18.77 & 0.7709 & \colorbox{yellow!25}{0.2721}
    & & 24.82 & 0.8355 & 0.2255
    & & 25.16 & 0.8566 & 0.1745
    & & 22.79 & 0.8235 & \colorbox{yellow!25}{0.2115} \\

    GaussianShader~\cite{jiang2023gaussianshader}
    & 14.55 & 0.6423 & 0.4955
    & & 19.97 & 0.7244 & 0.4507
    & & 22.58 & 0.7862 & 0.3057
    & & 19.70 & 0.7363 & 0.3819 \\

    GS-IR~\cite{liang2023gs}
    & 15.98 & 0.6969 & 0.4200
    & & 22.28 & 0.8088 & 0.3067
    & & 22.84 & 0.8212 & 0.2236
    & & 20.25 & 0.7796 & 0.2999 \\

    NeRF-DS~\cite{yan2023nerfdsneuralradiancefields}
    & 19.70 & 0.7813 & 0.2974
    & & 25.35 & \colorbox{orange!25}{0.8703} & 0.2552
    & & 24.99 & \colorbox{orange!25}{0.8705} & 0.2001
    & & 23.24 & \colorbox{yellow!25}{0.8384} & 0.2402 \\

    HyperNeRF~\cite{park2021hypernerf}
    & \colorbox{red!25}{21.22} & \colorbox{yellow!25}{0.7829} & 0.3166
    & & 16.54 & 0.8200 & 0.2810
    & & 19.92 & 0.8521 & 0.2142
    & & 19.01 & 0.8167 & 0.2615 \\

    EnvGS~\cite{xie2024envgs}
    & 15.33 & 0.6662 & 0.4005
    & & 21.84 & 0.8029 & 0.3032
    & & 23.74 & \colorbox{yellow!25}{0.8637} & 0.1922
    & & 20.21 & 0.7833 & 0.2837 \\

    SpectroMotion~\cite{fan2024spectromotion}
    & \colorbox{yellow!25}{20.84} & \colorbox{orange!25}{0.8172} & \colorbox{orange!25}{0.2198}
    & & \colorbox{orange!25}{26.49} & \colorbox{yellow!25}{0.8657} & \colorbox{orange!25}{0.1889}
    & & \colorbox{yellow!25}{25.22} & \colorbox{yellow!25}{0.8705} & \colorbox{orange!25}{0.1513}
    & & \colorbox{orange!25}{24.17} & \colorbox{orange!25}{0.8522} & \colorbox{orange!25}{0.1778} \\

    Ours
    & \colorbox{orange!25}{21.10} & \colorbox{red!25}{0.8415} & \colorbox{red!25}{0.1821}
    & & \colorbox{red!25}{27.39} & \colorbox{red!25}{0.9154} & \colorbox{red!25}{0.1559}
    & & \colorbox{red!25}{27.95} & \colorbox{red!25}{0.9178} & \colorbox{red!25}{0.1242}
    & & \colorbox{red!25}{24.91} & \colorbox{red!25}{0.8880} & \colorbox{red!25}{0.1471} \\

    \bottomrule
    \end{tabular}%
    }
\end{table*}

\subsection{Comparison with Baseline}
\noindent\textbf{Quantitative Comparison Results.} We compare our method with several state-of-the-art baselines on the NeRF-DS dataset, as shown in \autoref{tab:whole_scene_tab}. Among them, Deformable 3DGS \citep{yang2023deformablegs}, 4DGS \citep{yang2023gs4d}, and HyperNeRF \citep{park2021hypernerf} are designed for dynamic scene reconstruction; GaussianShader \citep{jiang2023gaussianshader}, GS-IR \citep{liang2023gs}, and EnvGS \citep{xie2024envgs} target static specular reconstruction; while NeRF-DS \citep{yan2023nerfdsneuralradiancefields} and SpectroMotion \citep{fan2024spectromotion} focus on dynamic specular scene reconstruction.

\begin{table}[t]
\centering
\small
\setlength{\tabcolsep}{2pt} %
\caption{\textbf{Quantitative comparison on HyperNeRF~\citep{park2021hypernerf}.} \colorbox{red!25}{Best} and \colorbox{orange!25}{second best} results are highlighted.}
\label{tab:hyper_main}
\vspace{-2mm}
\begin{tabular}{lccc}
\toprule
Method & PSNR↑ & SSIM↑ & LPIPS↓ \\
    \midrule
    \multicolumn{4}{c}{\textit{General dynamic reconstruction methods}} \\
    \midrule
Deformable 3DGS~\cite{yang2023deformablegs} &22.78 &0.6201 &{0.3380} \\
4DGS~\cite{yang2023gs4d} &{24.89} &{0.6781} &{0.3422} \\
HyperNeRF~\cite{park2021hypernerf} &{23.11} &{0.6387} &0.3968 \\
    \midrule
    \multicolumn{4}{c}{\textit{Specular reconstruction methods}} \\
    \midrule
NeRF-DS~\cite{yan2023nerfdsneuralradiancefields} &\colorbox{red!25}{23.65} &\colorbox{red!25}{0.6405} &0.3972 \\
SpectroMotion~\cite{fan2024spectromotion} &22.22 &0.6088 &\colorbox{orange!25}{0.3295} \\
GaussianShader~\cite{jiang2023gaussianshader} &18.55 &0.5452 &0.4795 \\
GS-IR~\cite{liang2023gs} &19.87 &0.5729 &0.4498 \\
Ours &\colorbox{orange!25}{22.47} &\colorbox{orange!25}{0.6328} &\colorbox{red!25}{0.3106} \\

\bottomrule
\end{tabular}
\vspace{-13pt}
\end{table}
We also evaluate our method on the HyperNeRF dataset, as shown in~\autoref{tab:hyper_main}, where it demonstrates competitive performance compared to state-of-the-art baselines.
Our method achieves superior performance, which we attribute to two key factors: first, it avoids approximation when computing reflection ray directions by relying on accurate surface normals; second, it incorporates a physically informed model of the specular imaging process. These two components together allow for sharper, more realistic specular detail reconstruction under complex dynamic conditions, leading to significant improvements in quantitative metrics.

\noindent\textbf{Qualitative Comparison Results.}
\autoref{fig:maincompare} presents qualitative comparisons with several state-of-the-art methods. We compare both dynamic scene reconstruction methods \citep{yang2023gs4d}, \citep{fan2024spectromotion} and static specular reconstruction methods \citep{jiang2023gaussianshader}. As shown, static methods such as  \citet{jiang2023gaussianshader}, which do not incorporate temporal consistency across frames, often suffer from severe artifacts in dynamic regions, including disappearance, blurriness, and ghosting, which significantly degrade the visual quality. Additionally, \citet{yang2023gs4d} explicitly models dynamic motion, but lacks consideration of specular components. As a result, it fails to capture sharp and detailed specular effects, leading to fragmented or missing details in highly reflective areas. As for \citet{fan2024spectromotion}, due to its inability to model near-field reflections, the apple reflected in the mirror is not reconstructed in the Press case, and artifacts appear in other cases as well. In contrast, our method produces visually coherent reconstructions with significantly sharper and more detailed specular reflections, effectively preserving both temporal consistency and high-frequency view-dependent effects.
\begin{figure*}[t!]
    \centering
    \includegraphics[width=\linewidth]{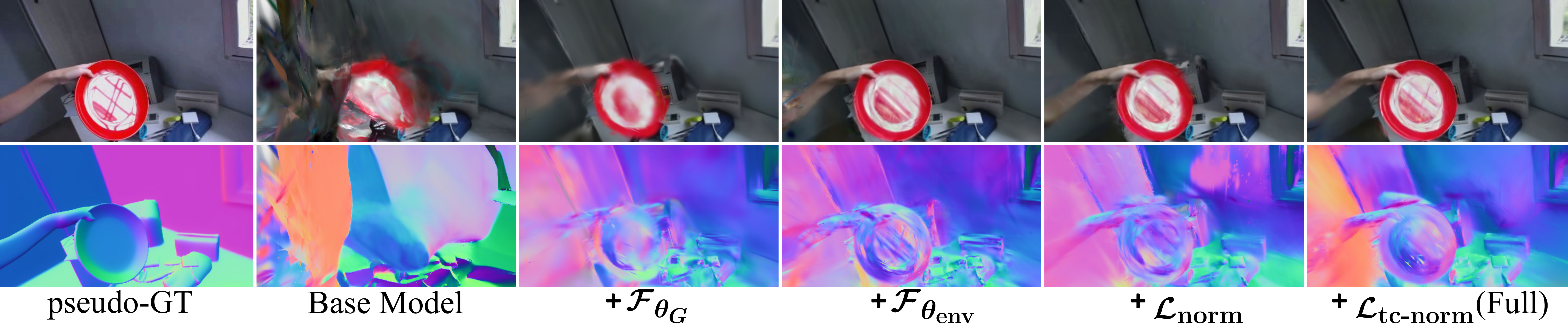}  %
    \caption{\textbf{Qualitative comparison of ablation study on different components.} "\texttt{+}" denotes the incremental addition of each component to the previous configuration, starting from the base model.}
    \label{fig:ablation}
\end{figure*}
\subsection{Ablation on Different Components.}
\begin{table}[t]
\centering
\small
\setlength{\tabcolsep}{5pt} %
\caption{\textbf{Ablation studies on different components.}}
\label{tab:abalation_with_losses}
\vspace{-2mm}
\begin{tabular}{cccc|ccc}
    \toprule
    $\mathcal{F}_{\theta_G}$ &  $\mathcal{F}_{\theta_{\text{env}}}$ & $\mathcal{L}_{\text{norm}}$ & $\mathcal{L}_{\text{tc-norm}}$ & PSNR↑ & SSIM↑ & LPIPS↓ \\
    \midrule
     & &  &  & 16.29 & 0.6655 & 0.4126 \\
    \checkmark & &  &  & 18.97 & 0.7793 & 0.2389 \\
    \checkmark &\checkmark  &  & & 20.25 & 0.8201 & 0.2313 \\
    \checkmark & \checkmark &\checkmark  & & 22.12 & 0.8441 & 0.2149 \\
    \checkmark & \checkmark & \checkmark & \checkmark & \textbf{25.23} & \textbf{0.8855} & \textbf{0.1536} \\
    \bottomrule
\end{tabular}
\vspace{-13pt}
\end{table}
We conduct ablation studies on four scenes (As, Press, Plate, and Bell) from the NeRF-DS \citep{yan2023nerfdsneuralradiancefields} dataset. The quantitative results in \autoref{tab:abalation_with_losses} report the average performance across these four scenes, while qualitative comparisons are shown in \autoref{fig:ablation}.

\noindent\textbf{Base Model.}
Our base model excludes the Time-Conditioned Residual Network $\mathcal{F}_{\theta_G}$, the residual correction network $\mathcal{F}_{\theta_{\text{env}}}$, Geometry-Aligned Normal Loss $\mathcal{L}_{\text{norm}}$, Temporal-Consistent Normal Supervision Loss $\mathcal{L}_{\text{tc-norm}}$. As shown in the first row of \autoref{tab:abalation_with_losses} and the “Base Model” column of \autoref{fig:ablation}, this configuration performs poorly due to the lack of dynamic modeling and geometric supervision. The results appear blurry and fail to recover scene structure, while the estimated normals are severely misaligned, indicating its inability to handle dynamic specular effects.

\noindent\textbf{+ Time-Conditioned Residual Network.}
We first add the Time-Conditioned Residual Network $\mathcal{F}_{\theta_G}$ to capture dynamic motion which yields notable improvements. The structure becomes more distinguishable, though specular regions remain blurry due to missing environment modeling and normal refinement.

\noindent\textbf{+ Residual Correction Network on Dynamic Environment.}
Adding the residual correction network $\mathcal{F}_{\theta_{\text{env}}}$ enables dynamic environment modeling which yields further improvements. Visually, specular regions become sharper and more realistic, normal maps capture finer geometric details.

\noindent\textbf{+ Geometry-Aligned Normal Loss.}
To improve geometry, we introduce the Geometry-Aligned Normal Loss $\mathcal{L}_{\text{norm}}$ which enhances surface normal and reflection direction accuracy, resulting in clearer specular regions.

\noindent\textbf{Full Model.}
Finally, we incorporate the Temporal-Consistent Normal Supervision Loss $\mathcal{L}_{\text{tc-norm}}$, which supplies temporally consistent pseudo ground-truth normals. The last row of \autoref{tab:abalation_with_losses} and the “+ $\mathcal{L}_{\text{tc-norm}}$ (Full)” column in \autoref{fig:ablation} show that this yields the best quantitative and qualitative performance, with improved normal consistency and sharper specular reflections across frames.

\section{Conclusion}

We presented \textit{ReflFlow}, a framework for dynamic specular scene reconstruction from monocular video. By combining Residual Material-Augmented 2DGS with Dynamic Environment Gaussians in a hybrid rendering pipeline combining rasterization and ray tracing, our method enables physically informed modeling of view-dependent reflections. Extensive experiments demonstrate that ReflFlow consistently outperforms prior approaches, particularly in challenging specular regions.

\section*{Acknowledgements}
This research is supported by NSF IIS-2525840, CNS-2432534, ECCS-2514574, NIH 1RF1MH133764-01 and Cisco Research unrestricted gift. This article solely reflects opinions and conclusions of authors and not funding agencies.

\section*{Impact Statement}
This paper presents work whose goal is to advance the field of 3D reconstruction and computer vision. Our method focuses on reconstructing dynamic specular scenes from monocular video, which has applications in AR/VR, visual effects, and robotics. We have conducted our research using publicly available datasets (NeRF-DS and HyperNeRF) with appropriate citations. Our method does not involve human subjects, private data collection, or raise immediate ethical concerns. While the technology could potentially be misused for creating deceptive visual content, we emphasize the importance of responsible deployment and recommend appropriate disclosure when synthetic content is generated using our method. We will release our code and trained models upon acceptance to facilitate reproduction and future research.

\bibliography{References}

\begin{thebibliography}{77}
\providecommand{\natexlab}[1]{#1}
\providecommand{\url}[1]{\texttt{#1}}
\expandafter\ifx\csname urlstyle\endcsname\relax
  \providecommand{\doi}[1]{doi: #1}\else
  \providecommand{\doi}{doi: \begingroup \urlstyle{rm}\Url}\fi

\bibitem[Bagdasarian et~al.(2025)Bagdasarian, Knoll, Li, Barthel, Hilsmann,
  Eisert, and Morgenstern]{bagdasarian2024zip}
Bagdasarian, M.~T., Knoll, P., Li, Y.-H., Barthel, F., Hilsmann, A., Eisert,
  P., and Morgenstern, W.
\newblock {3DGS.zip}: A survey on 3d {Gaussian} splatting compression methods.
\newblock \emph{Computer Graphics Forum}, 44\penalty0 (2), 2025.

\bibitem[Barron et~al.(2021)Barron, Mildenhall, Tancik, Hedman, Martin-Brualla,
  and Srinivasan]{mipnerf}
Barron, J.~T., Mildenhall, B., Tancik, M., Hedman, P., Martin-Brualla, R., and
  Srinivasan, P.~P.
\newblock Mip-nerf: A multiscale representation for anti-aliasing neural
  radiance fields, 2021.

\bibitem[Barron et~al.(2022)Barron, Mildenhall, Verbin, Srinivasan, and
  Hedman]{mipnerf360}
Barron, J.~T., Mildenhall, B., Verbin, D., Srinivasan, P.~P., and Hedman, P.
\newblock {Mip-NeRF} 360: Unbounded anti-aliased neural radiance fields.
\newblock In \emph{Proceedings of the IEEE/CVF Conference on Computer Vision
  and Pattern Recognition (CVPR)}, 2022.

\bibitem[Barron et~al.(2023)Barron, Mildenhall, Verbin, Srinivasan, and
  Hedman]{zipnerf}
Barron, J.~T., Mildenhall, B., Verbin, D., Srinivasan, P.~P., and Hedman, P.
\newblock Zip-nerf: Anti-aliased grid-based neural radiance fields.
\newblock \emph{ICCV}, 2023.

\bibitem[Bi et~al.(2024)Bi, Zeng, Zeng, Pei, Feng, Zhou, and Wu]{bi2024gs3}
Bi, Z., Zeng, Y., Zeng, C., Pei, F., Feng, X., Zhou, K., and Wu, H.
\newblock Gs3: Efficient relighting with triple gaussian splatting.
\newblock In \emph{SIGGRAPH Asia 2024 Conference Papers}, pp.\  1--12, 2024.

\bibitem[Bin et~al.(2025)Bin, Hu, Wang, Chen, and Wang]{normalcrafter}
Bin, Y., Hu, W., Wang, H., Chen, X., and Wang, B.
\newblock Normalcrafter: Learning temporally consistent normals from video
  diffusion priors, 2025.
\newblock URL \url{https://arxiv.org/abs/2504.11427}.

\bibitem[Burley(2012)]{burley2012disney}
Burley, B.
\newblock Physically-based shading at disney.
\newblock In \emph{ACM SIGGRAPH 2012 Courses}, pp.\  1--7, 2012.

\bibitem[Cao \& Johnson(2023)Cao and Johnson]{Cao2023HEXPLANE}
Cao, A. and Johnson, J.
\newblock Hexplane: A fast representation for dynamic scenes.
\newblock \emph{CVPR}, 2023.

\bibitem[Chen et~al.(2022)Chen, Xu, Geiger, Yu, and Su]{tensorf}
Chen, A., Xu, Z., Geiger, A., Yu, J., and Su, H.
\newblock Tensorf: Tensorial radiance fields.
\newblock In \emph{European Conference on Computer Vision (ECCV)}, 2022.

\bibitem[Chen et~al.(2025)Chen, Li, Ye, Wang, Xie, Zhai, Wang, Liu, Bao, and
  Zhang]{chen2024pgsr}
Chen, D., Li, H., Ye, W., Wang, Y., Xie, W., Zhai, S., Wang, N., Liu, H., Bao,
  H., and Zhang, G.
\newblock {PGSR}: Planar-based {Gaussian} splatting for efficient and
  high-fidelity surface reconstruction.
\newblock \emph{IEEE Transactions on Visualization and Computer Graphics},
  31\penalty0 (9):\penalty0 6100--6111, 2025.

\bibitem[Chen et~al.(2024{\natexlab{a}})Chen, He, He, and Zhang]{chen2024pisr}
Chen, G., He, Y., He, L., and Zhang, H.
\newblock Pisr: Polarimetric neural implicit surface reconstruction for
  textureless and specular objects.
\newblock In \emph{Proceedings of the European Conference on Computer Vision
  (ECCV)}, 2024{\natexlab{a}}.

\bibitem[Chen et~al.(2024{\natexlab{b}})Chen, Wei, Li, Huang, Wang, and
  Lee]{chen2024vcr}
Chen, H., Wei, F., Li, C., Huang, T., Wang, Y., and Lee, G.~H.
\newblock {VCR-GauS}: View consistent depth-normal regularizer for {Gaussian}
  surface reconstruction.
\newblock In \emph{Advances in Neural Information Processing Systems
  (NeurIPS)}, volume~37, pp.\  139725--139750, 2024{\natexlab{b}}.

\bibitem[Chen et~al.(2024{\natexlab{c}})Chen, Chan, Shiu, Yen, Yeh, and
  Liu]{chen2024narcan}
Chen, T.-H., Chan, J.~W., Shiu, H.-S., Yen, S.-H., Yeh, C., and Liu, Y.-L.
\newblock Narcan: Natural refined canonical image with integration of diffusion
  prior for video editing.
\newblock In \emph{Advances in Neural Information Processing Systems
  (NeurIPS)}, 2024{\natexlab{c}}.

\bibitem[Fan et~al.(2024)Fan, Chang, Liu, Lee, Huang, Tseng, and
  Liu]{fan2024spectromotion}
Fan, C.-D., Chang, C.-W., Liu, Y.-R., Lee, J.-Y., Huang, J.-L., Tseng, Y.-C.,
  and Liu, Y.-L.
\newblock Spectromotion: Dynamic 3d reconstruction of specular scenes.
\newblock \emph{arXiv}, 2024.

\bibitem[Gao et~al.(2024)Gao, Planche, Zheng, Choudhuri, Chen, and
  Wu]{gao20246dgs}
Gao, Z., Planche, B., Zheng, M., Choudhuri, A., Chen, T., and Wu, Z.
\newblock 6dgs: Enhanced direction-aware gaussian splatting for volumetric
  rendering, 2024.
\newblock URL \url{https://arxiv.org/abs/2410.04974}.

\bibitem[Gao et~al.(2025)Gao, Planche, Zheng, Choudhuri, Chen, and
  Wu]{gao20257dgsunifiedspatialtemporalangulargaussian}
Gao, Z., Planche, B., Zheng, M., Choudhuri, A., Chen, T., and Wu, Z.
\newblock 7dgs: Unified spatial-temporal-angular gaussian splatting, 2025.
\newblock URL \url{https://arxiv.org/abs/2503.07946}.

\bibitem[Ge et~al.(2023)Ge, Hu, Zhao, Liu, and Chen]{ge2023refneus}
Ge, W., Hu, T., Zhao, H., Liu, S., and Chen, Y.-C.
\newblock Ref-neus: Ambiguity-reduced neural implicit surface learning for
  multi-view reconstruction with reflection, 2023.
\newblock Preprint.

\bibitem[Gu et~al.(2025)Gu, Wei, Zeng, Yao, and Zhang]{gu2024IRGS}
Gu, C., Wei, X., Zeng, Z., Yao, Y., and Zhang, L.
\newblock {IRGS}: Inter-reflective {Gaussian} splatting with 2d {Gaussian} ray
  tracing.
\newblock In \emph{Proceedings of the IEEE/CVF Conference on Computer Vision
  and Pattern Recognition (CVPR)}, pp.\  10943--10952, 2025.

\bibitem[Guo et~al.(2023)Guo, Sun, Dai, Chen, Ye, Tan, Ding, Zhang, and
  Wang]{guo2023forwardflow}
Guo, X., Sun, J., Dai, Y., Chen, G., Ye, X., Tan, X., Ding, E., Zhang, Y., and
  Wang, J.
\newblock Forward flow for novel view synthesis of dynamic scenes, 2023.
\newblock Preprint.

\bibitem[H{\"o}llein et~al.(2025)H{\"o}llein, Bo{\v{z}}i{\v{c}}, Zollh{\"o}fer,
  and Nie{\ss}ner]{hollein2024lm}
H{\"o}llein, L., Bo{\v{z}}i{\v{c}}, A., Zollh{\"o}fer, M., and Nie{\ss}ner, M.
\newblock {3DGS-LM}: Faster {Gaussian}-splatting optimization with
  {Levenberg-Marquardt}.
\newblock In \emph{Proceedings of the IEEE/CVF International Conference on
  Computer Vision (ICCV)}, pp.\  26740--26750, 2025.

\bibitem[Huang et~al.(2024{\natexlab{a}})Huang, Yu, Chen, Geiger, and
  Gao]{2dgs}
Huang, B., Yu, Z., Chen, A., Geiger, A., and Gao, S.
\newblock 2d gaussian splatting for geometrically accurate radiance fields.
\newblock In \emph{Special Interest Group on Computer Graphics and Interactive
  Techniques Conference Conference Papers ’24}, SIGGRAPH ’24, pp.\  1–11.
  ACM, July 2024{\natexlab{a}}.
\newblock \doi{10.1145/3641519.3657428}.
\newblock URL \url{http://dx.doi.org/10.1145/3641519.3657428}.

\bibitem[Huang et~al.(2024{\natexlab{b}})Huang, Sun, Yang, Lyu, Cao, and
  Qi]{huang2024scgs}
Huang, Y.-H., Sun, Y.-T., Yang, Z., Lyu, X., Cao, Y.-P., and Qi, X.
\newblock Sc-gs: Sparse-controlled gaussian splatting for editable dynamic
  scenes, 2024{\natexlab{b}}.
\newblock Preprint.

\bibitem[Jiang et~al.(2024)Jiang, Tu, Liu, Gao, Long, Wang, and
  Ma]{jiang2023gaussianshader}
Jiang, Y., Tu, J., Liu, Y., Gao, X., Long, X., Wang, W., and Ma, Y.
\newblock {GaussianShader}: 3d {Gaussian} splatting with shading functions for
  reflective surfaces.
\newblock In \emph{Proceedings of the IEEE/CVF Conference on Computer Vision
  and Pattern Recognition (CVPR)}, pp.\  5322--5332, 2024.

\bibitem[Kerbl et~al.(2023)Kerbl, Kopanas, Leimk{\"u}hler, and Drettakis]{3dgs}
Kerbl, B., Kopanas, G., Leimk{\"u}hler, T., and Drettakis, G.
\newblock 3d gaussian splatting for real-time radiance field rendering.
\newblock \emph{ACM Transactions on Graphics}, 42\penalty0 (4), July 2023.
\newblock URL \url{https://repo-sam.inria.fr/fungraph/3d-gaussian-splatting/}.

\bibitem[Kheradmand et~al.(2024)Kheradmand, Rebain, Sharma, Sun, Tseng, Isack,
  Kar, Tagliasacchi, and Yi]{kheradmand2024mcmcgs}
Kheradmand, S., Rebain, D., Sharma, G., Sun, W., Tseng, Y.-C., Isack, H., Kar,
  A., Tagliasacchi, A., and Yi, K.~M.
\newblock 3d {Gaussian} splatting as {Markov} chain {Monte} {Carlo}.
\newblock In \emph{Advances in Neural Information Processing Systems
  (NeurIPS)}, volume~37, pp.\  80965--80986, 2024.

\bibitem[Lee et~al.(2024)Lee, Rho, Sun, Ko, and Park]{compact3dgs}
Lee, J.~C., Rho, D., Sun, X., Ko, J.~H., and Park, E.
\newblock Compact 3d gaussian representation for radiance field.
\newblock In \emph{Proceedings of the IEEE/CVF Conference on Computer Vision
  and Pattern Recognition (CVPR)}, pp.\  21719--21728, 2024.

\bibitem[Li et~al.(2023{\natexlab{a}})Li, Tao, Yang, and Yang]{li2023human101}
Li, M., Tao, J., Yang, Z., and Yang, Y.
\newblock {Human101}: Training 100+ fps human {Gaussians} in 100s from 1 view.
\newblock \emph{arXiv preprint arXiv:2312.15258}, 2023{\natexlab{a}}.

\bibitem[Li et~al.(2025)Li, Pang, Fan, Huang, and Yang]{li2025tsgs}
Li, M., Pang, P., Fan, H., Huang, H., and Yang, Y.
\newblock {TSGS}: Improving {Gaussian} splatting for transparent surface
  reconstruction via normal and de-lighting priors.
\newblock In \emph{Proceedings of the 33rd ACM International Conference on
  Multimedia}, pp.\  7220--7229, 2025.

\bibitem[Li et~al.(2026)Li, Fan, and Yang]{li2026transnormal}
Li, M., Fan, H., and Yang, Y.
\newblock {TransNormal}: Dense visual semantics for diffusion-based transparent
  object normal estimation.
\newblock \emph{arXiv preprint arXiv:2602.00839}, 2026.

\bibitem[Li et~al.(2021)Li, Niklaus, Snavely, and Wang]{li2021nsff}
Li, Z., Niklaus, S., Snavely, N., and Wang, O.
\newblock Neural scene flow fields for space-time view synthesis of dynamic
  scenes.
\newblock In \emph{Proceedings of the IEEE/CVF Conference on Computer Vision
  and Pattern Recognition (CVPR)}, 2021.

\bibitem[Li et~al.(2023{\natexlab{b}})Li, Müller, Evans, Taylor, Unberath,
  Liu, and Lin]{li2023neuralangelo}
Li, Z., Müller, T., Evans, A., Taylor, R.~H., Unberath, M., Liu, M.-Y., and
  Lin, C.-H.
\newblock Neuralangelo: High-fidelity neural surface reconstruction.
\newblock In \emph{Proceedings of the IEEE/CVF Conference on Computer Vision
  and Pattern Recognition (CVPR)}, 2023{\natexlab{b}}.

\bibitem[Liang et~al.(2023{\natexlab{a}})Liang, Chen, Li, Chen, Panneer, and
  Vijaykumar]{liang2023envidr}
Liang, R., Chen, H., Li, C., Chen, F., Panneer, S., and Vijaykumar, N.
\newblock Envidr: Implicit differentiable renderer with neural environment
  lighting, 2023{\natexlab{a}}.
\newblock Preprint.

\bibitem[Liang et~al.(2023{\natexlab{b}})Liang, Zhang, Li, Yang, Guan, and
  Vijaykumar]{liang2023spidr}
Liang, R., Zhang, J., Li, H., Yang, C., Guan, Y., and Vijaykumar, N.
\newblock Spidr: Sdf-based neural point fields for illumination and
  deformation, 2023{\natexlab{b}}.
\newblock Preprint.

\bibitem[Liang et~al.(2023{\natexlab{c}})Liang, Khan, Li, Nguyen-Phuoc, Lanman,
  Tompkin, and Xiao]{liang2023gaufre}
Liang, Y., Khan, N., Li, Z., Nguyen-Phuoc, T., Lanman, D., Tompkin, J., and
  Xiao, L.
\newblock Gaufre: Gaussian deformation fields for real-time dynamic novel view
  synthesis, 2023{\natexlab{c}}.
\newblock Preprint.

\bibitem[Liang et~al.(2024)Liang, Zhang, Feng, Shan, and Jia]{liang2023gs}
Liang, Z., Zhang, Q., Feng, Y., Shan, Y., and Jia, K.
\newblock {GS-IR}: 3d {Gaussian} splatting for inverse rendering.
\newblock In \emph{Proceedings of the IEEE/CVF Conference on Computer Vision
  and Pattern Recognition (CVPR)}, pp.\  21644--21653, 2024.

\bibitem[Liu et~al.(2020)Liu, Gu, Lin, Chua, and Theobalt]{liu2020neural}
Liu, L., Gu, J., Lin, K.~Z., Chua, T.-S., and Theobalt, C.
\newblock Neural sparse voxel fields.
\newblock \emph{NeurIPS}, 2020.

\bibitem[Liu et~al.(2023{\natexlab{a}})Liu, Wang, Lin, Long, Wang, Liu, Komura,
  and Wang]{liu2023nero}
Liu, Y., Wang, P., Lin, C., Long, X., Wang, J., Liu, L., Komura, T., and Wang,
  W.
\newblock Nero: Neural geometry and brdf reconstruction of reflective objects
  from multiview images.
\newblock In \emph{Proceedings of SIGGRAPH}, 2023{\natexlab{a}}.

\bibitem[Liu et~al.(2023{\natexlab{b}})Liu, Gao, Meuleman, Tseng, Saraf, Kim,
  Chuang, Kopf, and Huang]{liu2023robustdynrf}
Liu, Y.-L., Gao, C., Meuleman, A., Tseng, H.-Y., Saraf, A., Kim, C., Chuang,
  Y.-Y., Kopf, J., and Huang, J.-B.
\newblock Robust dynamic radiance fields.
\newblock In \emph{Proceedings of the IEEE/CVF Conference on Computer Vision
  and Pattern Recognition (CVPR)}, 2023{\natexlab{b}}.

\bibitem[Lu et~al.(2024)Lu, Yu, Xu, Xiangli, Wang, Lin, and
  Dai]{lu2024scaffoldgs}
Lu, T., Yu, M., Xu, L., Xiangli, Y., Wang, L., Lin, D., and Dai, B.
\newblock Scaffold-gs: Structured 3d gaussians for view-adaptive rendering.
\newblock In \emph{Proceedings of the IEEE/CVF Conference on Computer Vision
  and Pattern Recognition (CVPR)}, pp.\  20654--20664, 2024.

\bibitem[Ma et~al.(2024)Ma, Liu, Wang, Liu, Liu, and Wang]{ma2024humannerfse}
Ma, C., Liu, Y.-L., Wang, Z., Liu, W., Liu, X., and Wang, Z.
\newblock Humannerf-se: A simple yet effective approach to animate humannerf
  with diverse poses.
\newblock In \emph{Proceedings of the IEEE/CVF Conference on Computer Vision
  and Pattern Recognition (CVPR)}, 2024.

\bibitem[Ma et~al.(2023)Ma, Agrawal, Turki, Kim, Gao, Sander, Zollhöfer, and
  Richardt]{ma2023specnerf}
Ma, L., Agrawal, V., Turki, H., Kim, C., Gao, C., Sander, P., Zollhöfer, M.,
  and Richardt, C.
\newblock Specnerf: Gaussian directional encoding for specular reflections,
  2023.

\bibitem[Mildenhall et~al.(2020)Mildenhall, Srinivasan, Tancik, Barron,
  Ramamoorthi, and Ng]{nerf}
Mildenhall, B., Srinivasan, P.~P., Tancik, M., Barron, J.~T., Ramamoorthi, R.,
  and Ng, R.
\newblock Nerf: Representing scenes as neural radiance fields for view
  synthesis.
\newblock In \emph{ECCV}, 2020.

\bibitem[M\"uller et~al.(2022)M\"uller, Evans, Schied, and
  Keller]{mueller2022instant}
M\"uller, T., Evans, A., Schied, C., and Keller, A.
\newblock Instant neural graphics primitives with a multiresolution hash
  encoding.
\newblock \emph{ACM Trans. Graph.}, 41\penalty0 (4):\penalty0 102:1--102:15,
  July 2022.
\newblock \doi{10.1145/3528223.3530127}.
\newblock URL \url{https://doi.org/10.1145/3528223.3530127}.

\bibitem[Park et~al.(2021{\natexlab{a}})Park, Sinha, Barron, Bouaziz, Goldman,
  Seitz, and Martin-Brualla]{park2021nerfies}
Park, K., Sinha, U., Barron, J.~T., Bouaziz, S., Goldman, D.~B., Seitz, S.~M.,
  and Martin-Brualla, R.
\newblock Nerfies: Deformable neural radiance fields.
\newblock In \emph{Proceedings of the IEEE/CVF International Conference on
  Computer Vision (ICCV)}, 2021{\natexlab{a}}.

\bibitem[Park et~al.(2021{\natexlab{b}})Park, Sinha, Hedman, Barron, Bouaziz,
  Goldman, Martin-Brualla, and Seitz]{park2021hypernerf}
Park, K., Sinha, U., Hedman, P., Barron, J.~T., Bouaziz, S., Goldman, D.~B.,
  Martin-Brualla, R., and Seitz, S.~M.
\newblock {HyperNeRF}: A higher-dimensional representation for topologically
  varying neural radiance fields.
\newblock \emph{ACM Transactions on Graphics}, 40\penalty0 (6):\penalty0
  238:1--238:12, 2021{\natexlab{b}}.

\bibitem[Pharr et~al.(2023)Pharr, Jakob, and Humphreys]{pbr}
Pharr, M., Jakob, W., and Humphreys, G.
\newblock \emph{Physically Based Rendering: From Theory to Implementation}.
\newblock The MIT Press, Cambridge, MA, USA, fourth edition, 2023.

\bibitem[Pumarola et~al.(2021)Pumarola, Corona, Pons-Moll, and
  Moreno-Noguer]{pumarola2020dnerf}
Pumarola, A., Corona, E., Pons-Moll, G., and Moreno-Noguer, F.
\newblock {D-NeRF}: Neural radiance fields for dynamic scenes.
\newblock In \emph{Proceedings of the IEEE/CVF Conference on Computer Vision
  and Pattern Recognition (CVPR)}, pp.\  10318--10327, 2021.

\bibitem[{Sara Fridovich-Keil and Alex Yu} et~al.(2022){Sara Fridovich-Keil and
  Alex Yu}, Tancik, Chen, Recht, and Kanazawa]{plenoxels}
{Sara Fridovich-Keil and Alex Yu}, Tancik, M., Chen, Q., Recht, B., and
  Kanazawa, A.
\newblock Plenoxels: Radiance fields without neural networks.
\newblock In \emph{CVPR}, 2022.

\bibitem[Stearns et~al.(2024)Stearns, Harley, Uy, Dubost, Tombari, Wetzstein,
  and Guibas]{stearns2024marbles}
Stearns, C., Harley, A.~W., Uy, M.~A., Dubost, F., Tombari, F., Wetzstein, G.,
  and Guibas, L.~J.
\newblock Dynamic {Gaussian} marbles for novel view synthesis of casual
  monocular videos.
\newblock In \emph{SIGGRAPH Asia 2024 Conference Papers}, pp.\  30:1--30:11.
  ACM, 2024.

\bibitem[Sun et~al.(2022)Sun, Sun, and Chen]{DVGO}
Sun, C., Sun, M., and Chen, H.
\newblock Direct voxel grid optimization: Super-fast convergence for radiance
  fields reconstruction.
\newblock In \emph{CVPR}, 2022.

\bibitem[Tang \& Cham(2024)Tang and Cham]{tang2024threeigs}
Tang, Z.~J. and Cham, T.-J.
\newblock {3iGS}: Factorised tensorial illumination for 3d {Gaussian}
  splatting.
\newblock In \emph{European Conference on Computer Vision (ECCV)}, Lecture
  Notes in Computer Science, pp.\  143--159. Springer, 2024.

\bibitem[Tao et~al.(2026)Tao, Planche, Nguyen, Wu, Liu, Wang, Gao, Zhang,
  Zheng, Wang, Choudhuri, Zhao, Kang, Chen, Yan, and
  Wu]{tao2026particlesfieldsreframingphoton}
Tao, J., Planche, B., Nguyen, V.~N., Wu, J., Liu, Y., Wang, H., Gao, Z., Zhang,
  G., Zheng, M., Wang, F., Choudhuri, A., Zhao, Z., Kang, W., Chen, T., Yan,
  Y., and Wu, Z.
\newblock From particles to fields: Reframing photon mapping with continuous
  {Gaussian} photon fields.
\newblock \emph{arXiv preprint arXiv:2512.12459}, 2026.

\bibitem[Tretschk et~al.(2021)Tretschk, Tewari, Golyanik, Zollh{\"o}fer,
  Lassner, and Theobalt]{tretschk2021nonrigid}
Tretschk, E., Tewari, A., Golyanik, V., Zollh{\"o}fer, M., Lassner, C., and
  Theobalt, C.
\newblock Nonrigid neural radiance fields: Reconstruction and novel view
  synthesis of a dynamic scene from monocular video.
\newblock In \emph{Proceedings of the IEEE/CVF International Conference on
  Computer Vision (ICCV)}, 2021.

\bibitem[Verbin et~al.(2022)Verbin, Hedman, Mildenhall, Zickler, Barron, and
  Srinivasan]{verbin2022refnerf}
Verbin, D., Hedman, P., Mildenhall, B., Zickler, T., Barron, J.~T., and
  Srinivasan, P.~P.
\newblock Ref-nerf: Structured view-dependent appearance for neural radiance
  fields.
\newblock In \emph{Proceedings of the IEEE/CVF Conference on Computer Vision
  and Pattern Recognition (CVPR)}, 2022.

\bibitem[Verbin et~al.(2024)Verbin, Srinivasan, Hedman, Mildenhall, Attal,
  Szeliski, and Barron]{verbin2024nerfcasting}
Verbin, D., Srinivasan, P.~P., Hedman, P., Mildenhall, B., Attal, B., Szeliski,
  R., and Barron, J.~T.
\newblock Nerf-casting: Improved view-dependent appearance with consistent
  reflections, 2024.
\newblock URL \url{https://arxiv.org/abs/2405.14871}.

\bibitem[Wang et~al.(2024)Wang, Rakotosaona, Niemeyer, Szeliski, Pollefeys, and
  Tombari]{wang2024unisdf}
Wang, F., Rakotosaona, M.-J., Niemeyer, M., Szeliski, R., Pollefeys, M., and
  Tombari, F.
\newblock Unisdf: Unifying neural representations for high-fidelity 3d
  reconstruction of complex scenes with reflections.
\newblock In \emph{Advances in Neural Information Processing Systems
  (NeurIPS)}, 2024.

\bibitem[Wang et~al.(2025{\natexlab{a}})Wang, Tao, Wu, Wang, Duan, Wang, Yang,
  and Yan]{wang2025xfieldphysicallygroundedrepresentation}
Wang, F., Tao, J., Wu, J., Wang, H., Duan, B., Wang, K., Yang, Z., and Yan, Y.
\newblock {X-Field}: A physically informed representation for 3d {X-ray}
  reconstruction.
\newblock In \emph{Advances in Neural Information Processing Systems
  (NeurIPS)}, volume~38, pp.\  150612--150637, 2025{\natexlab{a}}.

\bibitem[Wang et~al.(2021)Wang, Liu, Liu, Theobalt, Komura, and
  Wang]{wang2021neus}
Wang, P., Liu, L., Liu, Y., Theobalt, C., Komura, T., and Wang, W.
\newblock Neus: Learning neural implicit surfaces by volume rendering for
  multi-view reconstruction.
\newblock In \emph{Advances in Neural Information Processing Systems
  (NeurIPS)}, 2021.

\bibitem[Wang et~al.(2025{\natexlab{b}})Wang, Ye, Gao, Zeng, Austin, Li, and
  Kanazawa]{wang2024shapeofmotion}
Wang, Q., Ye, V., Gao, H., Zeng, W., Austin, J., Li, Z., and Kanazawa, A.
\newblock Shape of motion: 4d reconstruction from a single video.
\newblock In \emph{Proceedings of the IEEE/CVF International Conference on
  Computer Vision (ICCV)}, pp.\  9660--9672, 2025{\natexlab{b}}.

\bibitem[Wang et~al.(2023)Wang, Han, Habermann, Daniilidis, Theobalt, and
  Liu]{wang2023neus2}
Wang, Y., Han, Q., Habermann, M., Daniilidis, K., Theobalt, C., and Liu, L.
\newblock Neus2: Fast learning of neural implicit surfaces for multi-view
  reconstruction.
\newblock In \emph{Proceedings of the IEEE/CVF International Conference on
  Computer Vision (ICCV)}, 2023.

\bibitem[Wang et~al.(2004)Wang, Bovik, Sheikh, and Simoncelli]{ssim}
Wang, Z., Bovik, A.~C., Sheikh, H.~R., and Simoncelli, E.~P.
\newblock Image quality assessment: From error visibility to structural
  similarity.
\newblock \emph{IEEE Transactions on Image Processing}, 13\penalty0
  (4):\penalty0 600--612, 2004.
\newblock \doi{10.1109/TIP.2003.819861}.

\bibitem[Wu et~al.(2025)Wu, Chen, Hu, Wu, Chen, Chen, Su, Lee, and
  Liu]{wu2025denver}
Wu, C.-H., Chen, S.-H., Hu, C.-Y., Wu, H.-Y., Chen, K.-H., Chen, Y.-Y., Su,
  C.-H., Lee, C.-K., and Liu, Y.-L.
\newblock Denver: Deformable neural vessel representations for unsupervised
  video vessel segmentation.
\newblock In \emph{Proceedings of the IEEE/CVF Conference on Computer Vision
  and Pattern Recognition (CVPR)}, 2025.

\bibitem[Wu et~al.(2024)Wu, Yi, Fang, Xie, Zhang, Wei, Liu, Tian, and
  Wang]{wu2023fourdgs}
Wu, G., Yi, T., Fang, J., Xie, L., Zhang, X., Wei, W., Liu, W., Tian, Q., and
  Wang, X.
\newblock 4d {Gaussian} splatting for real-time dynamic scene rendering.
\newblock In \emph{Proceedings of the IEEE/CVF Conference on Computer Vision
  and Pattern Recognition (CVPR)}, pp.\  20310--20320, 2024.

\bibitem[Xian et~al.(2021)Xian, Huang, Kopf, and Kim]{xian2021stnerf}
Xian, W., Huang, J.-B., Kopf, J., and Kim, C.
\newblock Space-time neural irradiance fields for free-viewpoint video.
\newblock In \emph{Proceedings of the IEEE/CVF Conference on Computer Vision
  and Pattern Recognition (CVPR)}, pp.\  9421--9431, 2021.

\bibitem[Xie et~al.(2025)Xie, Chen, Xu, Xie, Jin, Shen, Peng, Bao, and
  Zhou]{xie2024envgs}
Xie, T., Chen, X., Xu, Z., Xie, Y., Jin, Y., Shen, Y., Peng, S., Bao, H., and
  Zhou, X.
\newblock {EnvGS}: Modeling view-dependent appearance with environment
  {Gaussian}.
\newblock In \emph{Proceedings of the IEEE/CVF Conference on Computer Vision
  and Pattern Recognition (CVPR)}, pp.\  5742--5751, 2025.

\bibitem[Yan et~al.(2023)Yan, Li, and Lee]{yan2023nerfdsneuralradiancefields}
Yan, Z., Li, C., and Lee, G.~H.
\newblock Nerf-ds: Neural radiance fields for dynamic specular objects, 2023.
\newblock URL \url{https://arxiv.org/abs/2303.14435}.

\bibitem[Yang et~al.(2024{\natexlab{a}})Yang, Gao, Sun, Huang, Lyu, Zhou, Jiao,
  Qi, and Jin]{yang2024specgaussian}
Yang, Z., Gao, X., Sun, Y., Huang, Y., Lyu, X., Zhou, W., Jiao, S., Qi, X., and
  Jin, X.
\newblock Spec-gaussian: Anisotropic view-dependent appearance for 3d gaussian
  splatting, 2024{\natexlab{a}}.
\newblock Preprint.

\bibitem[Yang et~al.(2024{\natexlab{b}})Yang, Gao, Zhou, Jiao, Zhang, and
  Jin]{yang2023deformablegs}
Yang, Z., Gao, X., Zhou, W., Jiao, S., Zhang, Y., and Jin, X.
\newblock Deformable 3d {Gaussians} for high-fidelity monocular dynamic scene
  reconstruction.
\newblock In \emph{Proceedings of the IEEE/CVF Conference on Computer Vision
  and Pattern Recognition (CVPR)}, pp.\  20331--20341, 2024{\natexlab{b}}.

\bibitem[Yang et~al.(2024{\natexlab{c}})Yang, Yang, Pan, and
  Zhang]{yang2023gs4d}
Yang, Z., Yang, H., Pan, Z., and Zhang, L.
\newblock Real-time photorealistic dynamic scene representation and rendering
  with 4d gaussian splatting.
\newblock In \emph{International Conference on Learning Representations
  (ICLR)}, 2024{\natexlab{c}}.

\bibitem[Yariv et~al.(2020)Yariv, Kasten, Moran, Galun, Atzmon, Ronen, and
  Lipman]{yariv2020multiview}
Yariv, L., Kasten, Y., Moran, D., Galun, M., Atzmon, M., Ronen, B., and Lipman,
  Y.
\newblock Multiview neural surface reconstruction by disentangling geometry and
  appearance.
\newblock In \emph{Advances in Neural Information Processing Systems
  (NeurIPS)}, volume~33, 2020.

\bibitem[Ye et~al.(2024)Ye, Hou, and Zhou]{ye2024gsdr}
Ye, K., Hou, Q., and Zhou, K.
\newblock 3d {Gaussian} splatting with deferred reflection.
\newblock In \emph{ACM SIGGRAPH 2024 Conference Papers}. ACM, 2024.
\newblock \doi{10.1145/3641519.3657456}.

\bibitem[Yu et~al.(2024)Yu, Chen, Huang, Sattler, and
  Geiger]{yu2024mipsplatting}
Yu, Z., Chen, A., Huang, B., Sattler, T., and Geiger, A.
\newblock Mip-splatting: Alias-free 3d gaussian splatting.
\newblock In \emph{Proceedings of the IEEE/CVF Conference on Computer Vision
  and Pattern Recognition (CVPR)}, 2024.

\bibitem[Zhang et~al.(2023)Zhang, Yao, Li, Liu, Fang, McKinnon, Tsin, and
  Quan]{zhang2023neilfpp}
Zhang, J., Yao, Y., Li, S., Liu, J., Fang, T., McKinnon, D., Tsin, Y., and
  Quan, L.
\newblock Neilf++: Inter-reflectable light fields for geometry and material
  estimation, 2023.
\newblock Preprint.

\bibitem[Zhang et~al.(2018)Zhang, Isola, Efros, Shechtman, and Wang]{lpips}
Zhang, R., Isola, P., Efros, A.~A., Shechtman, E., and Wang, O.
\newblock The unreasonable effectiveness of deep features as a perceptual
  metric, 2018.
\newblock URL \url{https://arxiv.org/abs/1801.03924}.

\bibitem[Zhou et~al.(2025)Zhou, Li, Yang, and Yang]{zhou2025dreamrenderer}
Zhou, D., Li, M., Yang, Z., and Yang, Y.
\newblock {DreamRenderer}: Taming multi-instance attribute control in
  large-scale text-to-image models.
\newblock In \emph{Proceedings of the IEEE/CVF International Conference on
  Computer Vision (ICCV)}, pp.\  16712--16722, 2025.

\bibitem[Zhu et~al.(2024)Zhu, Liang, Chang, Deng, Lu, Yang, Zhang, and
  Zhang]{motiongs}
Zhu, R., Liang, Y., Chang, H., Deng, J., Lu, J., Yang, W., Zhang, T., and
  Zhang, Y.
\newblock Motiongs: Exploring explicit motion guidance for deformable 3d
  gaussian splatting, 2024.
\newblock URL \url{https://arxiv.org/abs/2410.07707}.

\bibitem[Zhu et~al.(2025)Zhu, Wang, and Yang]{zhu2024gsror}
Zhu, Z.-L., Wang, B., and Yang, J.
\newblock {GS-ROR}$^2$: Bidirectional-guided {3DGS} and {SDF} for reflective
  object relighting and reconstruction.
\newblock \emph{arXiv preprint arXiv:2406.18544}, 2025.

\end{thebibliography}
\bibliographystyle{icml2026}

\newpage
\appendix
\onecolumn
\appendix

\section{Coarse-to-Fine Training Strategy}

As described in~\autoref{subsec:c2f}, we design a coarse-to-fine training strategy to stabilize optimization and promote physically meaningful decomposition of appearance. Although our method explicitly separates the final pixel color into diffuse and specular components, supervision is applied only to the final rendered color $\mathbf{C}$. As a result, neither $\mathbf{C}{\text{diffuse}}$ nor $\mathbf{C}{\text{specular}}$ receives direct ground-truth supervision, rendering the decomposition inherently ill-posed and prone to instability, particularly during early training. This situation is akin to pulling a cart together without knowing which direction to exert force—the effort exists, but the alignment is lacking. Without proper regularization, the network may converge to trivial or degenerate solutions that minimize the reconstruction loss but fail to produce physically meaningful or interpretable results.

To mitigate this issue, we adopt a staged coarse-to-fine training strategy comprising a total of 60,000 training steps, divided into three progressive phases:

\begin{itemize}
    \item \textbf{Phase 1: Diffuse-Only Training (0–9k steps).} We begin by training only the diffuse rendering branch, using RGB ground truth to supervise geometry and diffuse color reconstruction. This phase establishes a reliable geometric foundation and reduces component entanglement during the early optimization. With reasonable geometry in place, the computation of reflection ray directions becomes more reliable, preventing gradient instability and enabling the network to learn specular color more robustly in the subsequent phases.
    
    \item \textbf{Phase 2: Specular-Only Training (9k–15k steps).} Once the diffuse branch reaches a stable state, we freeze its parameters and enable optimization of the specular rendering branch. This allows the network to learn dynamic environment and to learn specular appearance from reflection rays, guided by the reconstructed geometry in Phase 1.
    
    \item \textbf{Phase 3: Joint Fine-Tuning (15k–60k steps).} Finally, we unfreeze both branches and jointly optimize the entire network. This step encourages coordinated learning of diffuse and specular components and enables the network to refine geometry, normals, and material properties in a physically coherent manner.
\end{itemize}

This training strategy effectively balances the learning of diffuse and specular components. Empirically, we find that such staged optimization not only improves convergence stability but also enhances final rendering quality—producing sharper specular highlights and more accurate diffuse shading in dynamic scenes.

\begin{figure}[h!]
    \centering
    \includegraphics[width=\linewidth]{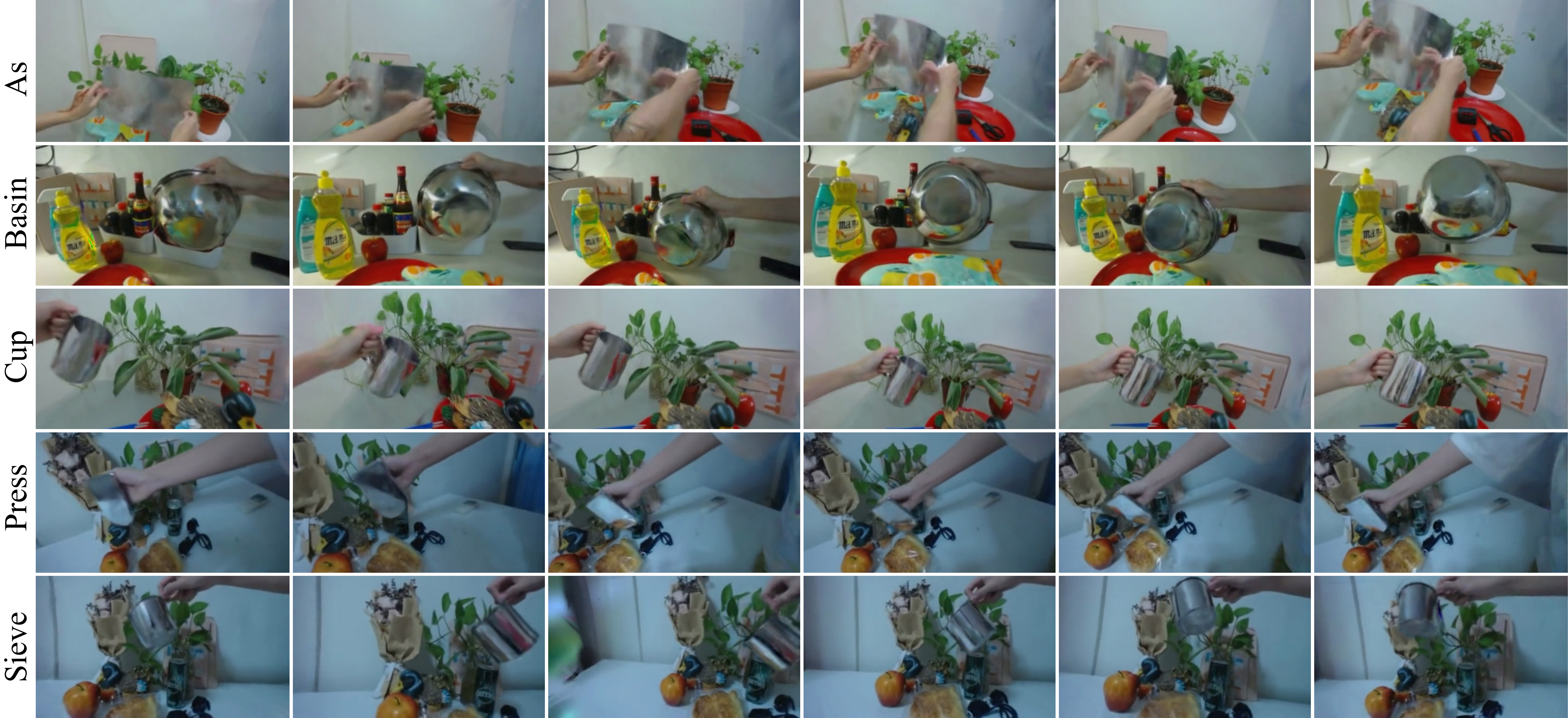}  %
    \caption{\textbf{More results on NeRF-DS datasets.} Our method can recover fine-grained specular details in dynamic specular reconstruction.}
    \label{fig:moreresults}
    \vspace{-15pt}
\end{figure}

\begin{figure}[h!]
    \centering
    \includegraphics[width=\linewidth]{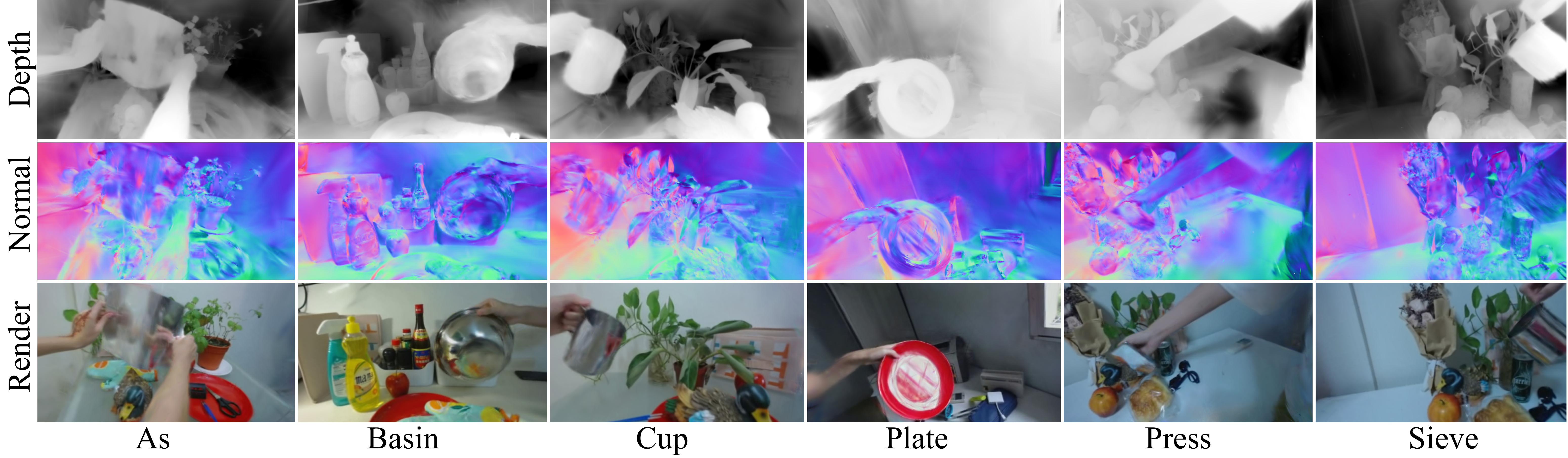}  %
    \caption{\textbf{Visualized our rendering images, normal maps, and depth maps.}}
    \label{fig:normal}
    \vspace{-12pt}
\end{figure}

\section{Datasets}
We evaluate our method on two datasets:

\begin{itemize}[leftmargin=10pt,topsep=0pt,itemsep=1pt,partopsep=1pt,parsep=1pt]
    \item \textbf{NeRF-DS \citep{yan2023nerfdsneuralradiancefields}}: A monocular video benchmark comprising seven real-world scenes with moving or deforming specular objects. We use the dataset's provided \texttt{points.npy} as the initial point cloud for our reconstruction. As shown in~\autoref{tab:whole_scene_tab} and~\autoref{fig:maincompare}, our method significantly outperforms existing baselines in both reconstruction accuracy and rendering quality on these challenging dynamic scenes.

    \item \textbf{HyperNeRF \citep{park2021hypernerf}}: A dataset of dynamic real-world scenes without a focus on specularity. We use the dataset's provided \texttt{points.npy} as the initial point cloud. We include it to evaluate generalization beyond specular-centric scenarios. As shown in~\autoref{tab:hyper_main}, our method achieves competitive performance, demonstrating its robustness in general dynamic scenes.
\end{itemize}

\section{Evaluation Metrics}
We evaluate our method using three image quality metrics: Peak Signal-to-Noise Ratio (PSNR), Structural Similarity Index (SSIM) \citep{ssim}, and LPIPS \citep{lpips}.

\section{Efficiency Comparison}
\begin{table}[h]
\centering
\small
\caption{\textbf{Efficiency comparison with SpectroMotion on NVIDIA RTX 6000 Ada.} Our method achieves comparable inference FPS while providing superior reconstruction quality.}
\label{tab:efficiency}
\vspace{+2mm}
\begin{tabular}{lcccc}
\toprule
Method & GPU & Iterations & Training Time & FPS↑ \\
\midrule
SpectroMotion~\citep{fan2024spectromotion} & RTX 6000 Ada & 40,000 & 1.1 hours & 33 \\
Ours & RTX 6000 Ada & 60,000 & 2.8 hours & 32 \\
\bottomrule
\end{tabular}
\end{table}

\section{Ablation for Coarse-to-Fine Training}

The coarse-to-fine schedule is designed to address an inherent optimization instability in dynamic specular reconstruction: early-stage supervision is dominated by high-frequency specular errors, which misguides geometry and normal estimation when reflection directions are still inaccurate. This leads to noisy normals, flickering specular components, and suboptimal convergence.

To validate its effectiveness, we added an ablation comparing training \textbf{with vs. without} the coarse-to-fine scheme. As shown in~\autoref{tab:c2f_ablation}, removing the coarse-to-fine strategy results in a significant drop in reconstruction quality (from 21.10 $\rightarrow$ 15.28 PSNR, 0.8415 $\rightarrow$ 0.5782 SSIM, 0.1821 $\rightarrow$ 0.4471 LPIPS).

These results demonstrate that the coarse-to-fine schedule is not merely a procedural choice but crucial for stable optimization and for obtaining physically meaningful diffuse/specular decomposition.

\begin{table}[t]
\centering
\small
\setlength{\tabcolsep}{5pt}
\caption{\textbf{Ablation for coarse-to-fine training.}}
\label{tab:c2f_ablation}
\vspace{-2mm}
\begin{tabular}{c|ccc}
\toprule
w/ coarse-to-fine & PSNR↑ & SSIM↑ & LPIPS↓ \\
\midrule
 & 15.28 & 0.5782 & 0.4471 \\
\checkmark & 21.10 & 0.8415 & 0.1821 \\
\bottomrule
\end{tabular}
\vspace{-10pt}
\end{table}

\section{More Results}
We present additional visual results in~\autoref{fig:moreresults} and~\autoref{fig:normal}. In~\autoref{fig:moreresults}, we show dynamic specular reconstructions over time. The results demonstrate that our method effectively recovers detailed specular highlights and maintains temporal consistency across frames. In~\autoref{fig:normal}, we visualize the depth maps, normal maps, and corresponding novel view renderings. These results indicate that our method produces high-quality geometry, which enables more accurate reflection ray direction estimation and ultimately leads to superior dynamic specular rendering.

\section{Limitation}

While \textit{ReflFlow} achieves high-quality dynamic specular reconstruction, its performance remains fundamentally constrained by the inherent challenges of monocular dynamic 3D reconstruction. Reconstructing accurate and temporally consistent surface geometry from a single moving camera is an ill-posed problem, particularly in complex dynamic scenes with fine-grained motions and non-rigid deformations. This geometric ambiguity directly impacts the reliability of reflection directions and surface normals, which are essential for high-fidelity specular rendering. Additionally, our method focuses on single-bounce reflections, which is the only physically identifiable case under monocular observation. Multi-bounce inter-reflections introduce additional ambiguities that cannot be reliably resolved without multi-view observations or strong priors. Similarly, glossy materials with spatially-varying BRDFs remain challenging due to the under-constrained nature of the problem—accurately decomposing material properties and lighting from monocular input requires complementary observations or additional supervision that are not available in our setup.

\end{document}